\documentclass{article}

\usepackage[preprint]{neurips_2025}

\usepackage[utf8]{inputenc} % allow utf-8 input
\usepackage[T1]{fontenc}    % use 8-bit T1 fonts
\usepackage{hyperref}       % hyperlinks
\usepackage{url}            % simple URL typesetting
\usepackage{booktabs}       % professional-quality tables
\usepackage{amsfonts}       % blackboard math symbols
\usepackage{nicefrac}       % compact symbols for 1/2, etc.
\usepackage{microtype}      % microtypography
\usepackage[table]{xcolor}         % colors

\usepackage{graphicx}
\usepackage{amsmath,amssymb,amsfonts}
\usepackage{dsfont}
\usepackage{subcaption}
\usepackage{caption}
\usepackage{multirow}
\usepackage{float}
\usepackage{amsthm}
\newtheoremstyle{definitionitalic}
  {5pt}   % Space above
  {5pt}   % Space below
  {\itshape}  % Body font
  {}      % Indent amount
  {\bfseries} % Theorem head font
  {.}     % Punctuation after theorem head
  {.5em}  % Space after theorem head
  {}      % Theorem head spec (can be left empty, meaning 'normal')

\theoremstyle{definitionitalic}
\newtheorem{definition}{Definition}
\usepackage{colortbl}  % needed for rule coloring
\definecolor{lightgray}{gray}{0.6} % custom grey for mild lines
\makeatother

\title{Factual Self-Awareness in Language Models: Representation, Robustness, and Scaling}

\author{%
  Hovhannes Tamoyan, Subhabrata Dutta, \textnormal{and} Iryna Gurevych \\
  Ubiquitous Knowledge Processing Lab (UKP Lab) \\
  Department of Computer Science and Hessian Center for AI (hessian.AI) \\
  Technical University of Darmstadt \\
  \texttt{www.ukp.tu-darmstadt.de}
}

\begin{document}

\maketitle

\begin{abstract}
    Factual incorrectness in generated content is one of the primary concerns in ubiquitous deployment of large language models (LLMs). Prior findings suggest LLMs can (sometimes) detect factual incorrectness in their generated content (i.e., fact-checking post-generation). In this work, we provide evidence supporting the presence of LLMs' internal compass that dictate the correctness of factual recall at the time of generation.
    We demonstrate that for a given subject entity and a relation, LLMs internally encode linear features in the Transformer's residual stream that dictate whether it will be able to recall the correct attribute (that forms a valid entity-relation-attribute triplet).
    This self-awareness signal is robust to minor formatting variations. 
    We investigate the effects of context perturbation via different example selection strategies.
    Scaling experiments across model sizes and training dynamics highlight that self-awareness emerges rapidly during training and peaks in intermediate layers.
    These findings uncover intrinsic self-monitoring capabilities within LLMs, contributing to their interpretability and reliability.\footnote{\href{https://github.com/UKPLab/arxiv2025-self-awareness}{https://github.com/UKPLab/arxiv2025-self-awareness}}
\end{abstract}

\section{Introduction}

With the recent advents in the ability of Large Language Models (LLMs), security concerns associated with LLM-based AI agents have increased in direct proportion~\citep{Huang2024}.
A lion's share of the security concerns about everyday LLM usage comes from their tendency to spit out made up facts---
a tendency mainly addressed under the broad description of {\em hallucination}, bearing an insinuation of {\em forgetfulness} of the models~\citep{10.1145/3703155}.
Recent research seeking to address hallucination has posed the question of transparency: is there any generalizable signal that can inform the presence (or absence) of certain knowledge in the internal representation of the model? 

Prior research in this direction follows two distinct lines.
\citet{DBLP:journals/corr/abs-2207-05221} showed that language models (LMs) can (most of the time) fact-check their output.
Multiple subsequent studies~\citep{Truth-I, Truth-II, Truth-III} have investigated and found that the LM encodes a notion of truth (and false) as linear directions within its representations, and these directions causally elicit the internal fact-checking.
This is similar to the broader line of research embodied in the self-reflection paradigm~\citep{DBLP:conf/nips/MadaanTGHGW0DPY23, DBLP:journals/corr/abs-self-reflexion}: let the model generate first and ask it to check itself. Recent literature has challenged the paradigm itself in problems such as reasoning and planning~\citep{DBLP:conf/iclr/StechlyVK25}. 

Another line of investigation reveals that language models (LMs) can demarcate between known and unknown entities~\citep{ferrando2024do}.
This demarcation, defined as the LMs ability (or inability) to recall at least three attributes about the candidate entity correctly, is also linearly represented within the residual stream of the Transformer model and instruction-tuned models re-purpose these linear features for refusal of questions related to unknown entities.
As opposed to the `truth direction'-style literature that analyzes the truthfulness of LMs on checking their own outputs, \citet{ferrando2024do} impose the notion of transparency at the time of generation.

\begin{figure}[!t]
    \centering
    \includegraphics[width=0.9\linewidth]{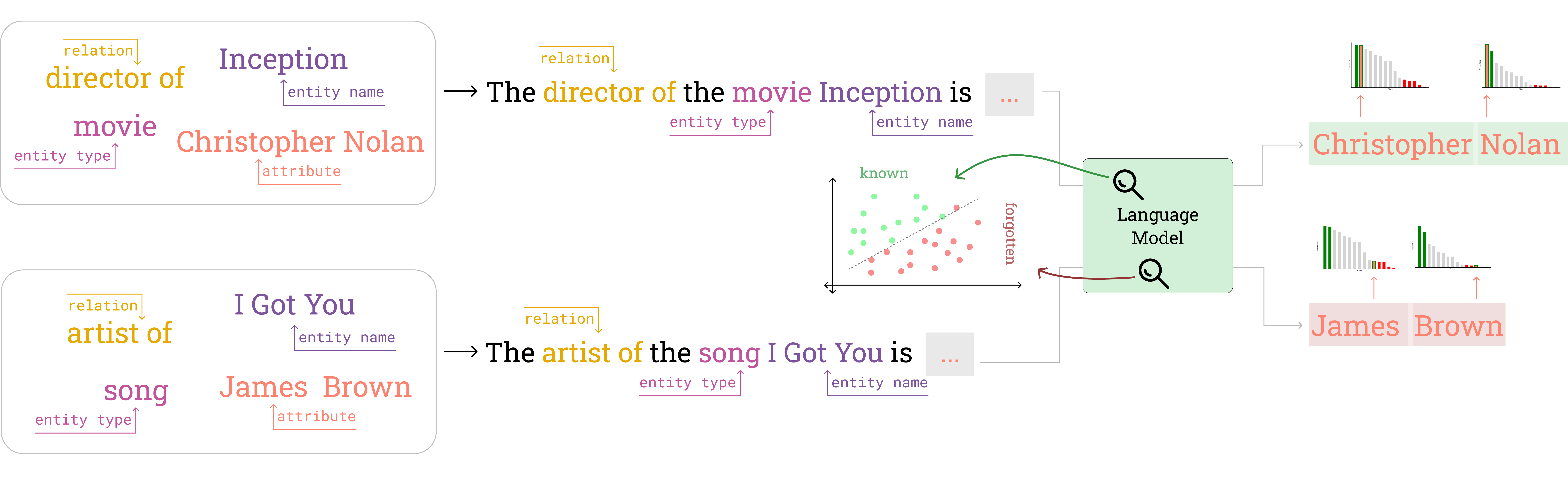}
    \caption{Given an input comprising entity type, entity name, and relation, we obtain the model's token-level prediction probabilities for the attribute. Tokens are labeled \texttt{known} if their gold label appears in the top-$k$ predictions, and \texttt{forgotten} if in the bottom-$l$. A sample is labeled \texttt{known} if it contains more \texttt{known} than \texttt{forgotten} tokens, and vice versa. Probabilities are visualized with color-coded bars: green (top-$k$), red (bottom-$l$), and gray (others). For example, ``Christopher Nolan'' falls in the top-$k$, labeling the sample as \texttt{known}, whereas ``James Brown'' appears in the bottom-$l$, labeling it as \texttt{forgotten}. Final token residuals are linearly probed to detect factual self-awareness.}
    \label{fig:method_ill}
\end{figure}
However, factual hallucination is not associated only with novel entities.
An LM may incorrectly recall a specific attribute of an entity while accurately recalling another (e.g., in the dataset we prepared, 1865 out of 3669 unique entities are neither completely known nor completely forgotten; Gemma-2 2B~\citep{team2024gemma} can recall at least one attribute about them correctly while erring on at least one attribute). Such a hallucination is intuitively harder to detect, compared to known-unknown entities in line with \citet{ferrando2024do}: an unknown entity (potentially unseen in the training data) is associated with a hitherto unseen lexical combination of a few successive tokens, whereas an unknown (or forgotten) factual association needs to be understood in a much deeper connection between entities and relationships.
In this work, we show that LMs encode meta-knowledge (i.e., ability or inability to correctly recall) about the fine-grained factual relationships as linear directions (see Figure~\ref{fig:method_ill}).
These directions are activated {\em before} generating the correct (or incorrect) factual recall, as opposed to the truth directions that are triggered after the generation is complete and the model is asked to check the generation.
This internal separability between correct and incorrect generations, which we denote as {\em known} and {\em forgotten} factual associations, is surprisingly robust to context perturbation.
We further investigate the effects of training and parameter counts on the factual self-awareness of the model; while a minimum model size is required to start encoding the signal, we find that it does so very early in training.

{\bf Contribution.} Towards investigating factual self-awareness of LMs at generation time, we construct a factual recall dataset. We investigate using Gemma-2 models (2B and 9B)~\citep{team2024gemma}, and the Pythia scaling suite~\citep{biderman2023pythia}, and propose a model-dependent annotation of known-forgotten facts based on logit distribution. We show that LMs construct linear subspaces within internal representations that can demarcate between an upcoming correct/incorrect recall (as opposed to faithfulness in post-hoc checking of correct/incorrect facts). The effects of context and prompt formatting on the formation of these linear subspaces of self-awareness are investigated. Finally, we demonstrate the appearance and improvement of factual self-awareness across two directions of LM scaling: amount of training for next token prediction and model parameters.

\section{Background and Related Work}
\label{sec:related_work}

In this section, we review existing literature dissecting language models' (LMs) self-awareness and provide background on linear probes and sparse autoencoders (SAEs), including their prior use in investigating truthfulness and self-awareness in LMs.

{\bf Self-awareness of LMs.} Prior efforts to make LMs transparent about their mistakes have primarily focused on mitigating hallucinations.
A common approach is to ask LMs unanswerable questions and test their ability to refrain~\citep{DBLP:conf/acl/YinSGWQH23, DBLP:conf/emnlp/BajpaiCD024}.
However, these questions are unanswerable not due to unknown or forgotten factual associations.
\citet{betley2025tell} similarly studied behavioral self-awareness in LMs.
By contrast, our work specifically targets factual self-awareness.
Prior work in this area follows the 'self-reflexion'~\citep{DBLP:conf/nips/MadaanTGHGW0DPY23,DBLP:journals/corr/abs-self-reflexion} paradigm: ask the model to reflect on its generation and assess its correctness.
\citet{DBLP:journals/corr/abs-2207-05221} supported LM self-awareness by demonstrating the success of such reflection on both multiple-choice and open-ended questions.
They also introduced fine-tuning strategies to calibrate model-generated scores of knowledge uncertainty.
\citet{DBLP:conf/nips/calibration-tuning} extended this 'teaching to be self-aware' paradigm via calibration tuning so that model-generated logits better reflect internal uncertainty.

{\bf Truthfulness and hallucination in internal representation.} 
Several works~\citep{Truth-I, Truth-II, Truth-III} have investigated how truthfulness is represented in model internals (i.e., intermediate layer outputs) in post-generation self-checking setups similar to \citet{DBLP:journals/corr/abs-2207-05221}.
The viability of self-consistency~\citep{DBLP:conf/acl/ZhangSWPWZ024, DBLP:conf/iclr/0002WSLCNCZ23} as a proxy for self-awareness has also been explored.
\citet{DBLP:conf/iclr/0026L0GWTFY24} demonstrate an implicit assumption of self-awareness via self-consistency: among a population of diverse generations, certain spectral patterns of internal states signal hallucination.
These methods are primarily designed for reasoning-based tasks, where multiple generations can lead to the same answer, in contrast to immediate factual recall.
\citet{DBLP:journals/corr/abs-2407-03282} analyzed training data to study unseen queries and found that LMs linearly represent seen vs. unseen queries in hidden states.
In a related direction, \citet{ferrando2024do} examined how LMs internally demarcate known from unknown entities.
They identified linearly encoded features that trigger when the model is queried about an unknown entity—features repurposed in chat-tuned models to elicit refusal.
A key limitation, however, lies in defining knowledge at the entity level: when an LM fabricates factual associations, it may invent attributes for a known entity.

{\bf Linear and sparse probing.} Linear probes are arguably the simplest lens for examining high-dimensional neural representations—given a labeled dataset of an expected behavior, a linear classifier is trained and tested on neural representations to detect whether the behavior is encoded.
Sparse autoencoders (SAEs) \citet{bricken2023towards, huben2023sparse}, by contrast, have recently gained popularity for uncovering interpretable decompositions of model latent representations without supervised data.
Both approaches align with the linear representation hypothesis~\citet{park2023linear, mikolov2013distributed}, which posits that interpretable features—such as sentiment or truthfulness—are embedded as linear directions within the representation space, and that model representations consist of sparse linear combinations of these directions~\cite{li2023inference, zou2023representation}.
While SAEs eliminate the need for supervision, they introduce the challenge of costly training: they must be trained on large volumes of data (and corresponding activations) to avoid data bias~\citep{kissane2024saes, DBLP:journals/corr/abs-2501-16496}.
SAE dimensionality strongly influences the contextual scale of interpretation~\citep{DBLP:journals/corr/abs-2503-17547}.
In this work, we first show the empirical similarity between linear probes and SAEs in locating factual self-awareness, then adopt linear probes for scalability.

\section{Experimental Setup}
\label{sec:experimental_setup}

{\bf Dataset.}
We construct a factual recall dataset covering four categories: football players, movies, cities, and songs, following the approach of \citet{ferrando2024do}.
We start with 1,000 entities for each of the following four categories: football player, movie, city, and song, limiting ourselves to a maximum of 10 relationships per entity.
For each entity we scrape associated features from Wikidata \citet{wikidata2023}.
Subsequently, we manually construct templates from the triplets\footnote{Factual associations are typically represented as triplets of the form ({\em entity name}, {\em relation}, {\em attribute}); we additionally use entity type to avoid ambiguities arising from shared naming across entities, e.g., computer scientist \textit{Michael Jordan} vs sportsman \textit{Michael Jordan}.} (\emph{entity type}, \emph{entity name}, \emph{relation}) to generate statements and predict the corresponding \emph{attribute}, as illustrated in \autoref{fig:method_ill}.

Due to the web-scale pretraining data used in training the Transformer-based LMs (as well as the non-availability of training data in case of most open-weight models), it is non-trivial to demarcate which factual associations are known (or unknown) to the LM.
We use a proxy definition for the same, using the logit distribution of the LM itself: if a model is able to signal that it can (or cannot) recall a certain attribute of an entity correctly (i.e., assign a high logit value to the respective token), we diagnose the behavior as {\em factually self-aware}.

We feed the samples (entity-relation pairs) into an LM to obtain the probability distribution over the predicted tokens, and classify the factual relations as either \texttt{known} or \texttt{forgotten}.
Since the gold labels are sequences of tokens (e.g. ``Christopher'', ``Nolan'') we generate the same number of tokens as in the gold label sequence for each sample and check how many of these appear in the top-$k$ or bottom $l$-th percentile of the model's output space~\footnote{Note that in this definition, $k$ is an integer denoting the count of tokens, while $l$ is a fraction denoting a subset of the vocabulary. This demarcation arises from the actual count of tokens in these bands: high probability tokens are exponentially fewer than low probability ones.}.
A sample is labeled as \texttt{known} if more of its gold label tokens appear among the top-$k$ predictions in the logit space.
Conversely, if more of the gold label tokens fall below the $l$-th percentile of the logit distribution, the sample is classified as \texttt{forgotten}.
This design choice avoids decoding and string matching errors by relying solely on model's output space.

We construct $(\text{entity type}, \text{entity name}, \text{relation})$ triplets using various templates, but some templates introduced spurious correlations due to their phrasing.
Details and examples of all templates are provided in \autoref{tab:input_templates} and \autoref{appendix:data_template}.
For subsequent experiments, we employ a template devoid of such artifacts, encompassing four relations per entity type and excluding problematic attributes (e.g., city coordinates). The complete list of relations is provided in \autoref{appendix:data_template}.

We construct the final dataset by setting top-$k = 500$ tokens and selecting the bottom-$l = 0.3$ fraction of the vocabulary space.
Detailed experiments on the impact of different $(k, l)$ pairs on factual self-awareness signals are presented in \autoref{subsec:k-l-impact}.
This procedure yields a total of 7,380 \texttt{known} and 7,268 \texttt{forgotten} samples for Gemma 2 2B~\citet{team2024gemma}.
The distribution of labels across entity types is provided in \autoref{tab:data_sample_distribution_2b} in \autoref{appendix:dataset_distribution}.
We partition the dataset $\mathcal{D}$ into training and test subsets with a $0{,}7$/$0{,}3$ split for subsequent experiments.

{\bf Linear Probe.}
Let ${\cal D} = \{(T_i, y_i)\}_{i=1}^N$ denote the labeled dataset, where each $T_i$ is a token sequence (e.g., a factual recall statement) and $y_i \in \{0, 1\}$ indicates the presence or absence of the feature (e.g., \texttt{known}/\texttt{forgotten}).
We run the model on each $T_i \in \cal{D}$ and extract the residual stream of the final token of the prompt template $x_{T_i}$, following \citet{meng2022locating, geva-etal-2023-dissecting, nanda2023fact}.

\begin{definition}
\label{def:probe}
For a residual stream representation ${x}_{l, T_i} \in \mathbb{R}^d$ of sample $T_i$ at layer $l$, a probe is a learnable function $f: \mathbb{R}^d \rightarrow \mathbb{R}$ trained to predict $y_i$ from $ x_{l, T_i}$.
\end{definition}

The linear probe serves as a simple diagnostic classifier that maps the residual stream output to a scalar output via a learned weight vector $\mathbf{w} \in \mathbb{R}^d$ and a bias term $b \in \mathbb{R}$.
At layer $l$, we introduce a linear probe and its corresponding optimization objective as:

\begin{equation}
f_l : \mathbb{R}^d \to \mathbb{R},\quad
f_l(x_{l,T_i}) = \mathbf{w}^\top x_{l,T_i} + b,
\qquad
\min_{\mathbf{w}, b}
\sum_{(T_i, y_i)\in \mathcal{D}}
\mathrm{BCE}\bigl(y_i,\sigma\bigl(f_l(x_{l,T_i})\bigr)\bigr).
\end{equation}
where $\sigma$ is the sigmoid function and $\mathrm{BCE}$ the binary cross‐entropy loss.
The parameters $\mathbf{w}$ and $b$ are learned by minimizing the binary cross‐entropy loss over the dataset $\mathcal{D}$.

To break symmetry and introduce controlled variation across layers, we initialize scalar biases as $b = 0.1 \times (-1)^l$, where the layer index $l$ deterministically seeds randomness via $\texttt{seed} + 100 \times l$ with $seed$\footnote{All experiments are conducted with three random seeds (73, 5, 120); we observe negligible variance and omit the results for brevity.}. This ensures consistent yet diverse initialization across layers.
To further encourage diversity in learned solutions, each probe is assigned a slightly different learning rate, scaled for layer $l$ as $\text{lr}_l = \texttt{base\_lr} \times \left(1.1 - 0.2 \cdot \frac{l}{L}\right)$, where \texttt{base\_lr} is the initial rate and $L$ the total number of layers. This encourages probes at different depths to converge to distinct solutions. All probes are optimized using Adam, with initial learning rate $1\mathrm{e}{-4}$ and weight decay $1\mathrm{e}{-5}$.

{\bf Separation Scores.}
Sparse Autoencoders (SAEs) conform to the definition of a probe as stated in Definition~\ref{def:probe}.
We collectively refer to the SAE-encoded representations and the outputs of the linear probe as \emph{activations}.
Following \citet{ferrando2024do}, we compute separation scores. For each latent dimension $j$ of the activation vector (with $j = 1$ for a linear probe), we calculate the proportion of instances with positive activations (i.e., greater than zero) separately for the \texttt{known} and \texttt{forgotten} sets: $g_{l,j}^{\text{known}} = \frac{\sum_i^{N^{\text{known}}} \mathds{1}\left[ a_{l,j}(x_{l,T_i}^{\text{known}}) > 0 \right]}{N^{\text{known}}}$ and $g_{l,j}^{\text{forgotten}} = \frac{\sum_i^{N^{\text{forgotten}}} \mathds{1}\left[ a_{l,j}(x_{l,T_i}^{\text{forgotten}}) > 0 \right]}{N^{\text{forgotten}}}$, where $N^{\text{known}}$ and $N^{\text{forgotten}}$ denote the total number of prompts, and $x_{l,T_i}^{\text{known}}$ and $x_{l,T_i}^{\text{forgotten}}$ represent the latent activations for the \texttt{known} and \texttt{forgotten} samples, respectively, in each subset of $\mathcal{D}$.
Latent separation scores (or vectors) are computed as the difference between these proportions: $s_{l,j}^{\text{known}} = g_{l,j}^{\text{known}} - g_{l,j}^{\text{forgotten}}$ and $s_{l,j}^{\text{forgotten}} = g_{l,j}^{\text{forgotten}} - g_{l,j}^{\text{known}}$, where $s_{l}^{\text{known}}$ is used to detect \texttt{known} entities, and $s_{l}^{\text{forgotten}}$ is used to detect \texttt{forgotten} entities.

\textbf{Computational Resource Requirements.}
All experiments are conducted on NVIDIA A100-SXM4-80GB GPUs.
Models with less than 7B parameters are run on a single GPU, while larger models are executed on two GPUs to accommodate memory and computational requirements.

\section{Generation-time factual self-awareness in LMs}
\label{sec:generation_time_factual_self_awareness}

\subsection{Linear Probes vs. Sparse Autoencoders (SAEs)}
We utilize the Gemma 2 2B and 9B models from the Gemma Scope framework~\citep{lieberum2024gemma}, which provides a suite of SAEs pretrained on the activations of each layer of the Gemma 2 models~\citep{team2024gemma}.
We train linear probes on the residual stream hooks of these models and generate the corresponding separation plots on test sets for both the 2B and 9B variants using both SAE and linear probing methods.
We select the top-five (one for linear probe) latent dimensions with the highest separation scores from the \texttt{known} and \texttt{forgotten} vectors for each entity type $t$ and layer $l$.
To assess generality and robustness, we compute $\text{MaxMin}^{\text{known}, l} = \max_j \min_t s_{l,j}^{\text{known}, t}$ and analogously define $\text{MaxMin}^{\text{forgotten}, l}$, where $j$ indexes latent dimensions.

\begin{figure}[ht]
    \centering
    \includegraphics[width=0.48\linewidth]{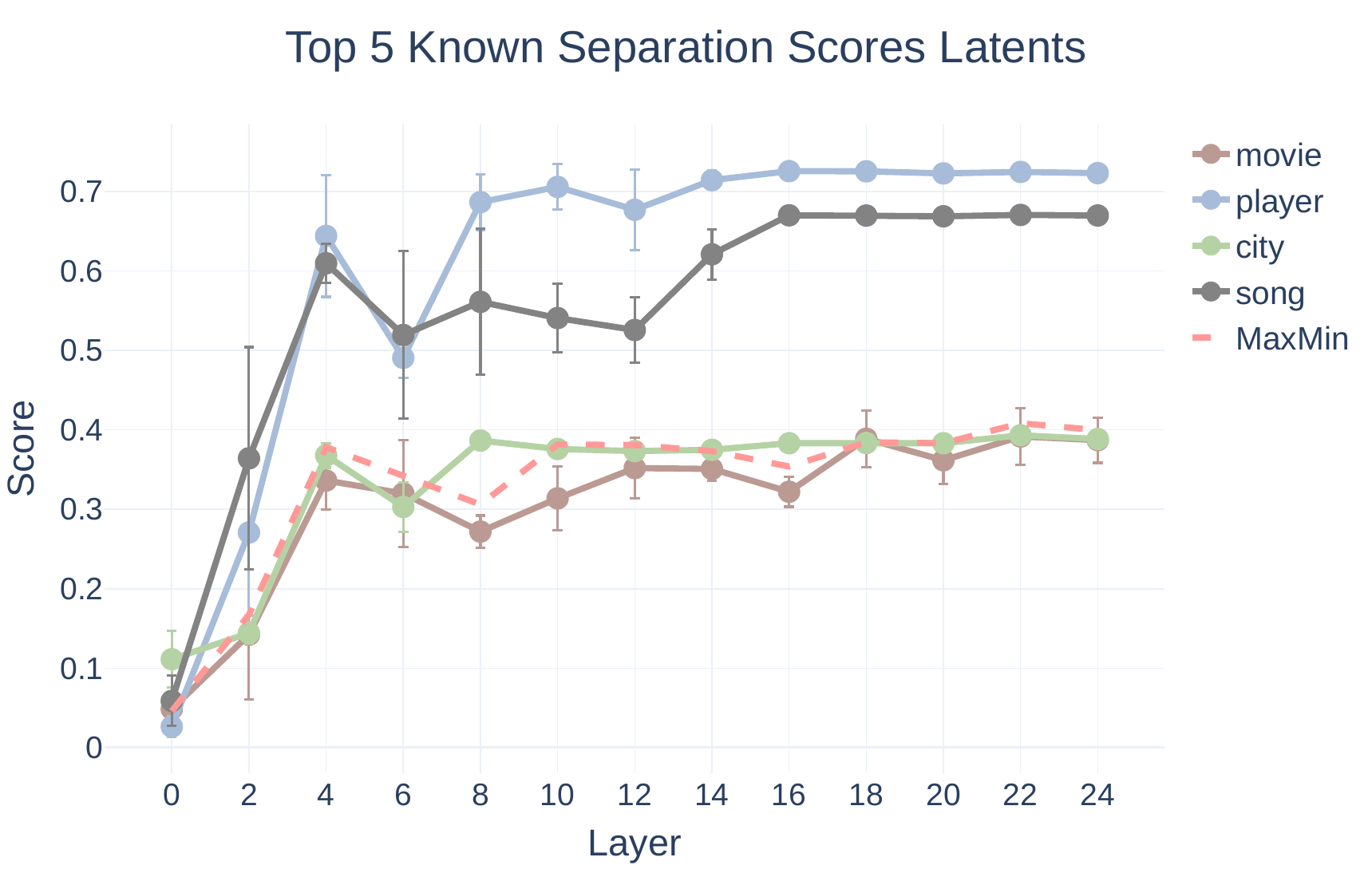}
    \hfill
    \includegraphics[width=0.48\linewidth]{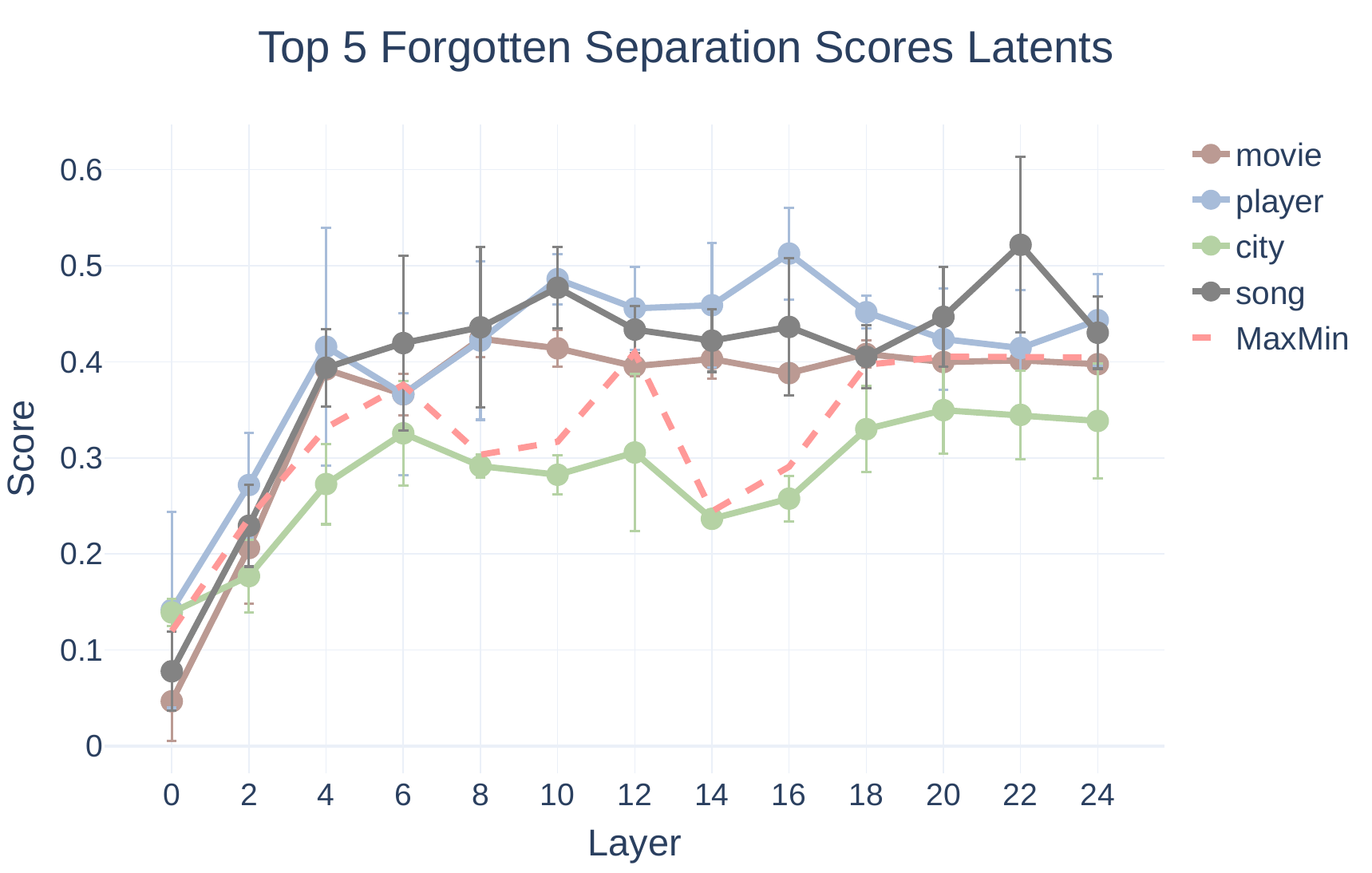}
    \caption{Top-five latent separation scores across transformer layers using SAE activations from Gemma 2 2B. Left: For \textbf{known} entities exhibit clear layer-wise separation, peaking around layers 6–14. Right: For \textbf{forgotten} entities, separation scores are lower and more variable, indicating reduced disentanglement. Categories include \texttt{movie}, \texttt{player}, \texttt{city}, and \texttt{song}; MaxMin denotes the difference between max and min class means.}
    \label{fig:sae_latent_separation_comparison_gemma_2b}
\end{figure}

\begin{figure}[ht]
    \centering
    \includegraphics[width=0.48\linewidth]{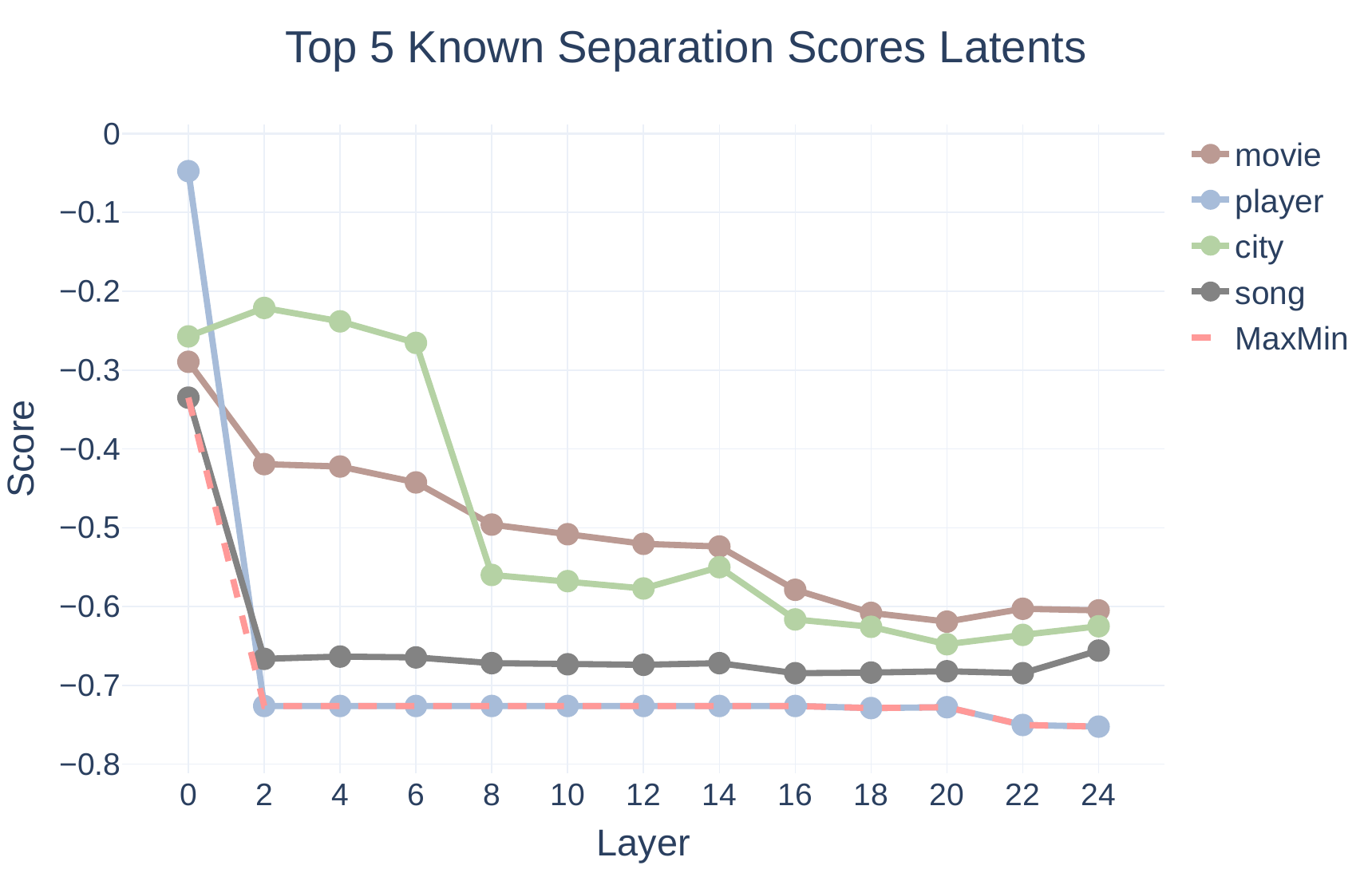}
    \hfill
    \includegraphics[width=0.48\linewidth]{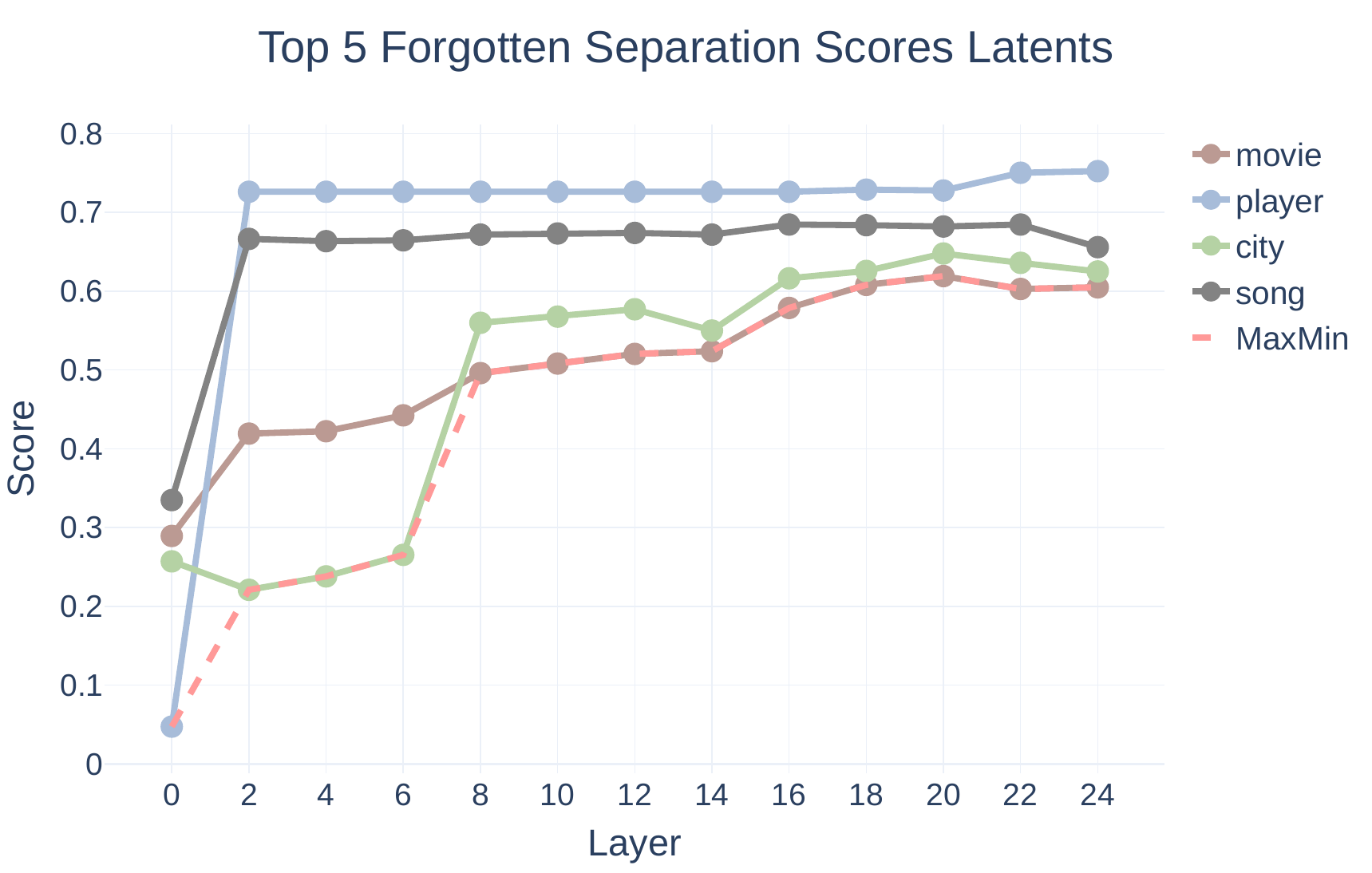}
    \caption{Latent separation scores across layers using Linear Probe activations from Gemma 2 2B. Left: \textbf{Known} entities show separation scores that are identical in magnitude but negated in sign compared to \textbf{forgotten} entities (right), indicating that the same latents are used for both but with reversed class-directional structure. Categories include \texttt{movie}, \texttt{player}, \texttt{city}, and \texttt{song}; MaxMin denotes the difference between maximum and minimum class means.}
    \label{fig:linear_prob_latent_separation_comparison_gemma_2b}
    \vspace{-4mm}
\end{figure}

We illustrate the evolution of separation scores across layers for SAE activations in \autoref{fig:sae_latent_separation_comparison_gemma_2b} (Gemma 2 9B model results are provided in \autoref{appendix:separation_scores}). As shown by the red curve, $\text{MaxMin}_{l}$ increases in the intermediate layers.
This pattern suggests that the most \emph{generalized} latents—those consistently separating \texttt{known} from \texttt{forgotten} entities across all types are primarily located in the middle and final layers.
Linear probe separation scores follow a similar trend, as shown in \autoref{fig:linear_prob_latent_separation_comparison_gemma_2b}; however, since these activations are scalar, the \texttt{known} and \texttt{forgotten} scores appear as mirror images, having equal magnitudes but opposite signs.

We train linear probes on the Gemma 2 (2B, 9B)~\citep{team2024gemma} and Pythia (70M, 1.4B, 6.9B, 12B)~\citep{biderman2023pythia} models, evaluating performance using standard binary classification metrics and reporting the accuracy improvement over a random baseline, denoted as $\Delta$.
Metrics from the final training epoch are reported for both the training and test subsets, as shown in \autoref{tab:probing_methods_comparison}.
Among all models, Gemma 2 2B exhibits the strongest performance, achieving the highest test set separation score with $\Delta = 0.311$.
Within the Pythia family, the 12B model performs best ($\Delta = 0.120$); however, a substantial performance gap remains relative to Gemma 2 2B.

\begin{table}[htbp]
\centering
\caption{
Linear probing results on Gemma 2 and Pythia model families. Metrics are reported on training and test subsets from the final epoch (3). Accuracy gains over random baselines are indicated by $\Delta$, with values in parentheses denoting standard deviations. Gemma 2 2B achieves the highest test set performance, while Pythia 12B is strongest within its family, though with a notable gap.
}
\label{tab:probing_methods_comparison}
\resizebox{\textwidth}{!}{%
\begin{tabular}{llcc|ccc}
\toprule
\textbf{Model} & \textbf{Subset} & \textbf{Loss} & \textbf{AUC ROC} 
& \multicolumn{3}{c}{\textbf{Accuracies}} \\
\cmidrule(lr){5-7}
& & & & \textbf{Observed} & \textbf{Random Baseline} & \textbf{$\Delta$ (Observed - Baseline)} \\
\midrule

\multirow{2}{*}{Gemma 2 2B} 
  & Train & \textbf{0.383} (0.063)& \textbf{0.901} (0.061)& \textbf{0.833} (0.052)& 0.501& \textbf{0.332}\\
  & Test  & 0.397 (0.064)& 0.896 (0.060)& 0.820 (0.056)& 0.509& \textbf{0.311}\\
\arrayrulecolor{darkgray}\cmidrule(lr){1-7}

\multirow{2}{*}{Gemma 2 9B} 
  & Train & 0.393 (0.052)& 0.899 (0.050)& 0.826 (0.040)& 0.555& 0.271\\
  & Test  & \textbf{0.387} (0.050)& \textbf{0.903} (0.047)& \textbf{0.829} (0.032)& 0.564& 0.265\\
\arrayrulecolor{black}\specialrule{0.1em}{2pt}{2pt} % thicker black separator

\multirow{2}{*}{Pythia 70M} 
  & Train & 0.473 (0.008)& 0.546 (0.029)& 0.818 (0.001)& 0.818& 0.000 \\
  & Test  & 0.465 (0.005)& 0.551 (0.022)& 0.822 (0.001)& 0.822& 0.000 \\
\arrayrulecolor{darkgray}\cmidrule(lr){1-7}

\multirow{2}{*}{Pythia 1.4B} 
  & Train & \textbf{0.358} (0.032)& 0.843 (0.057)& \textbf{0.837} (0.011)& 0.807& 0.030\\
  & Test  & \textbf{0.365} (0.031)& 0.842 (0.056)& \textbf{0.831} (0.013)& 0.803& 0.028\\
\cmidrule(lr){1-7}

\multirow{2}{*}{Pythia 6.9B} 
  & Train & 0.393 (0.028)& \textbf{0.852} (0.048)& 0.829 (0.014)& 0.747& 0.082\\
  & Test  & 0.395 (0.026)& \textbf{0.857} (0.047)& 0.827 (0.011)& 0.746& 0.081\\
\cmidrule(lr){1-7}

\multirow{2}{*}{Pythia 12B} 
  & Train & 0.442 (0.030)& 0.842 (0.046)& 0.798 (0.017)& 0.687& \textbf{0.111}\\
  & Test  & 0.464 (0.027)& 0.846 (0.043)& 0.794 (0.022)& 0.674& \textbf{0.120}\\
\arrayrulecolor{black}\bottomrule
\end{tabular}
}
\vspace{-4mm}
\end{table}

Beyond the overall advantage of the Gemma 2 2B model, several notable trends emerge from the probing results. Within each model family, increasing parameter count does not consistently improve linear probe performance.
For instance, while Gemma 2 9B achieves a slightly higher test AUC-ROC than Gemma 2 2B, its accuracy gain over the random baseline ($\Delta$) is lower.
This suggests that larger models do not necessarily produce more linearly decodable representations of self-awareness. A similar pattern holds for the Pythia models: although the 12B variant achieves the highest test-time $\Delta$, smaller versions like Pythia 6.9B and 1.4B show comparable accuracies, albeit with smaller gains over their baselines. Pythia 70M represents a degenerate case where accuracy matches the random baseline ($\Delta = 0$), indicating that the smallest model fails to encode self-awareness features.

\begin{figure}[ht]
\centering
\begin{subfigure}{0.49\linewidth}
    \includegraphics[width=\linewidth]{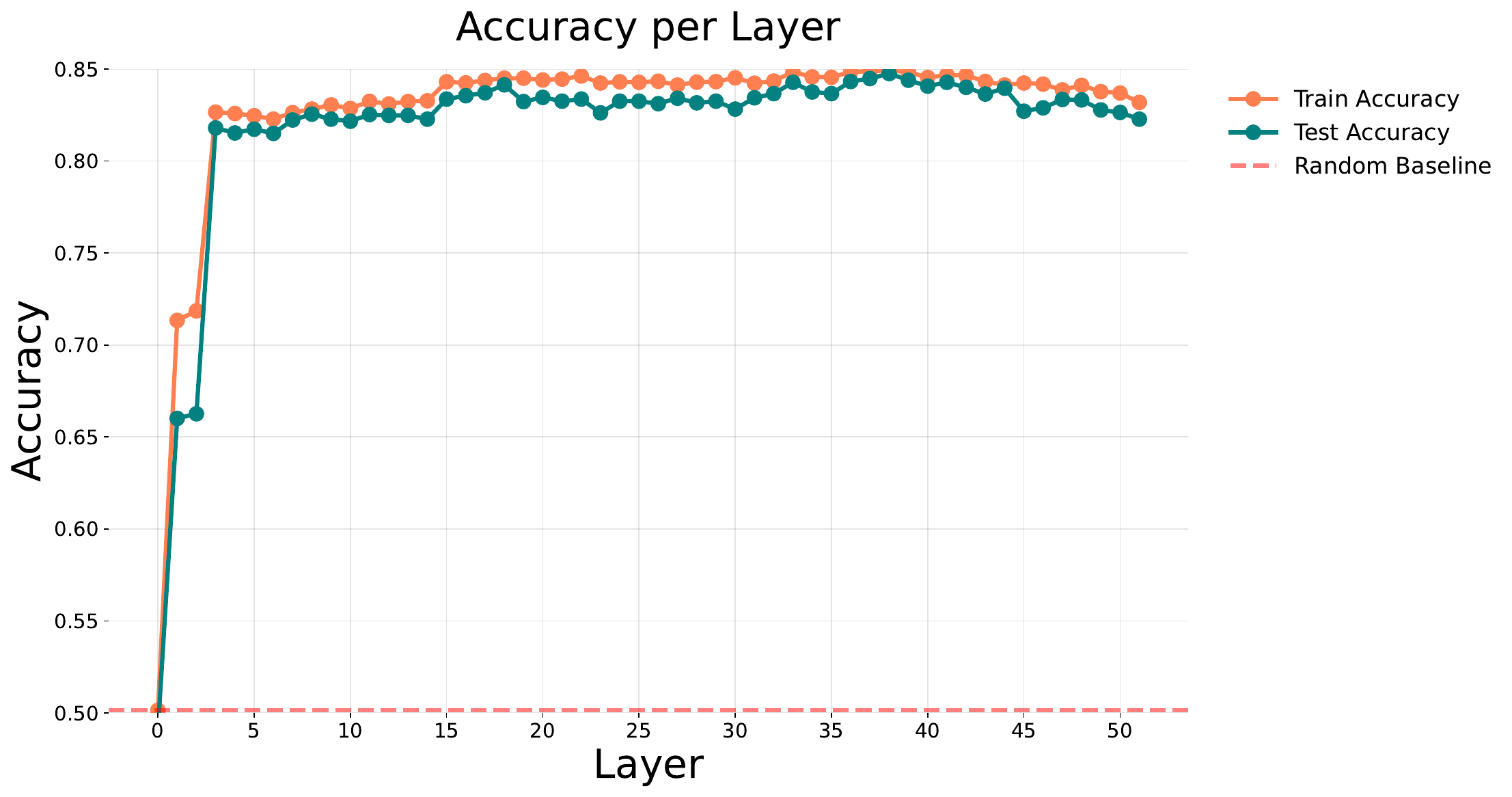}
    \caption{Gemma 2 2B}
    \label{fig:gemma-graph}
\end{subfigure}
\hfill
\begin{subfigure}{0.49\linewidth}
    \includegraphics[width=\linewidth]{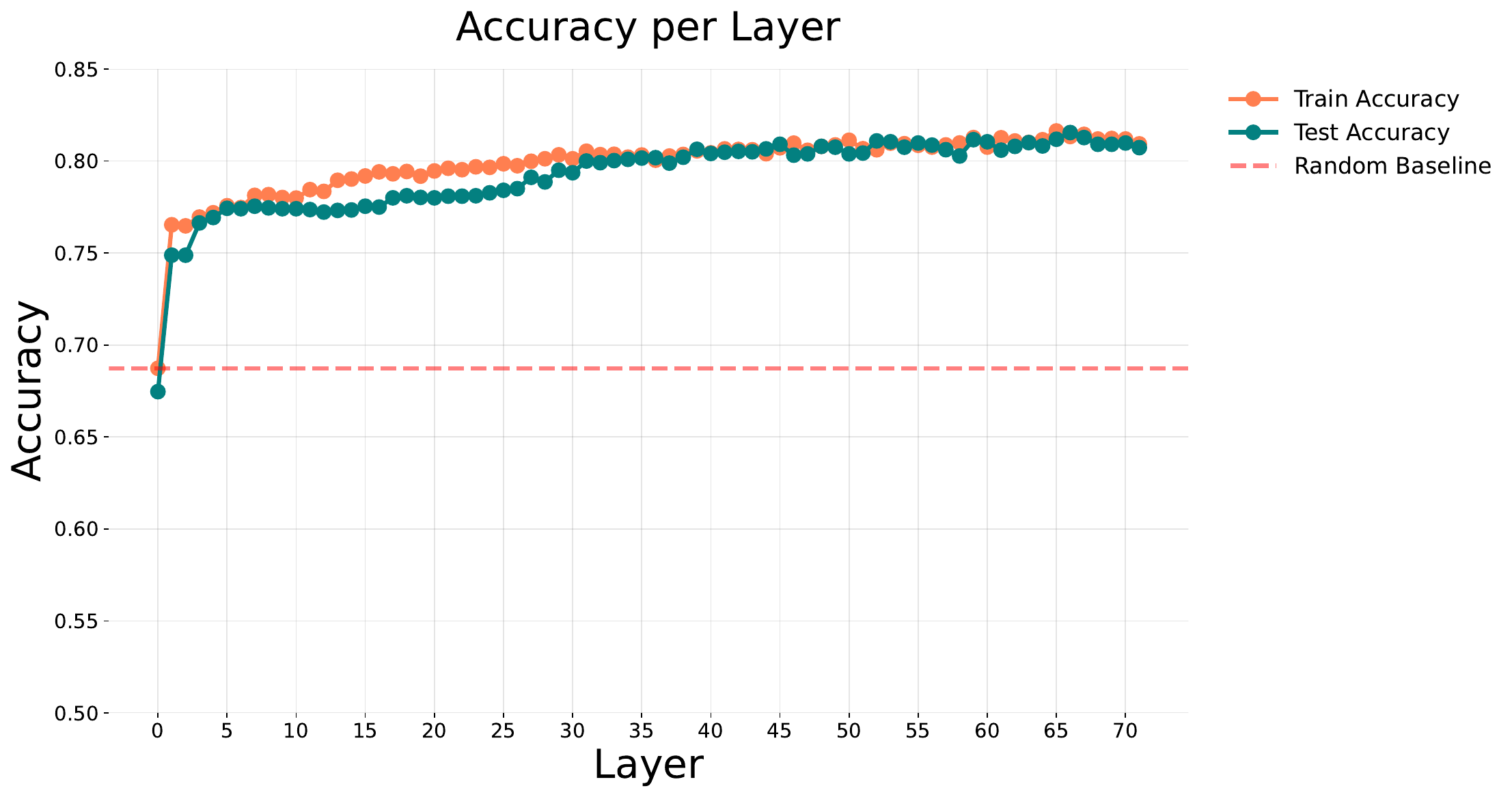}
    \caption{Pythia 12B}
    \label{fig:pythia-graph}
\end{subfigure}
\caption{Layer-wise linear probe accuracy for Gemma 2 2B and Pythia 12B. Orange/blue: train/test; red dashed: random baseline.}
\label{fig:graph-comparison}
\vspace{-4mm}
\end{figure}

We further examine the distribution of self-awareness signals across model layers by analyzing layer-wise linear probe accuracy, as shown in \autoref{fig:graph-comparison}.
For both Gemma 2 2B and Pythia 12B, accuracy rises sharply in the initial layers before stabilizing.
In Gemma 2 2B, performance plateaus around the fifth layer, reaching a peak test accuracy of approximately 0.82; well above the random baseline.
Accuracy remains consistently high in subsequent layers, with a slight decline in the final three layers, suggesting that self-awareness directions are preserved throughout the network depth.

In contrast, Pythia 12B shows a slower but steady accuracy increase across layers.
Its final test accuracy remains below that of Gemma 2 2B, consistent with earlier results.
Notably, the random baseline for Pythia 12B is substantially higher than for Gemma 2 2B (approximately 0.69 vs. 0.50).
These patterns suggest that Gemma 2 2B achieves linearly accessible self-awareness representations earlier and maintains them more robustly, while Pythia 12B requires deeper processing to approach similar performance.

\subsection{Robustness Against Context Perturbation}
\label{subsec:context_perturbations}

To what extent is a model's factual self-awareness robust to changes in input context?
To address this question, we train linear probes on each model layer and evaluate their performance on contextually modified input samples.
We design four targeted experiments to systematically assess the test-time robustness of self-awareness directions under such perturbations.

{\bf Quotation Marks.}
Enclose the entity name within single or double quotation marks:
\begin{center}
\vspace{-4mm}
    \includegraphics[width=0.6\linewidth]{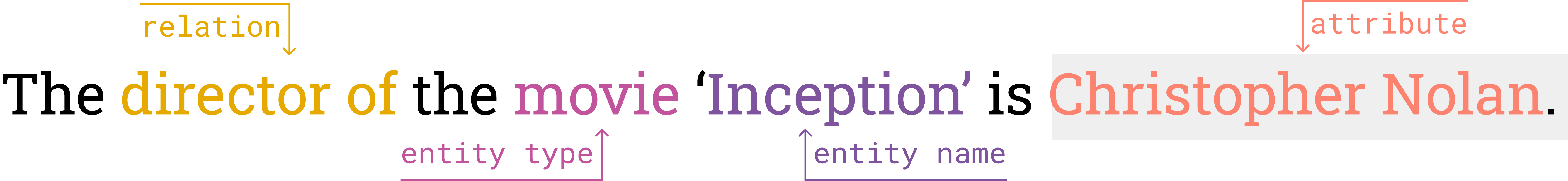}
    \vspace{-4mm}
\end{center}

Prompt formatting, including punctuation, spacing, and quoting, has been shown to significantly impact model performance \citep{gonen2023demystifying, sclar2024quantifying}.

{\bf Statement Question.}
Prepend a natural language question crafted from the sample quadruples:
\begin{center}
\vspace{-4mm}
    \includegraphics[width=0.95\linewidth]{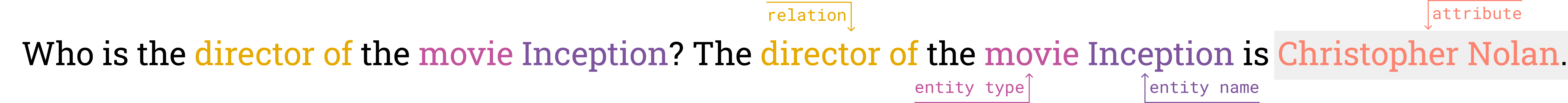}
    \vspace{-4mm}
\end{center}

Rephrasing inputs as questions can improve model reasoning and accuracy, with even minor phrasing changes affecting outputs and models performance \citep{kojima2022large, mizrahi2024state}.

{\bf Few-Shot.}
Prepend few-shot context by adding a small set of sample (e.g., three) from the dataset to the input.
These samples are chosen according to one of two entity modes. 

\textbf{Only}: all selected samples have the same entity name; however, the relation and attribute may vary:
\begin{center}
\vspace{-4mm}
    \includegraphics[width=.95\linewidth]{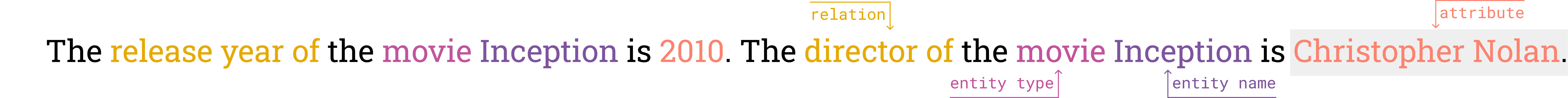}
    \vspace{-4mm}
\end{center}

\textbf{Unique}: each entity name appears only once in the context; however, the relations between entities are not necessarily distinct.
\begin{center}
\vspace{-4mm}
    \includegraphics[width=.95\linewidth]{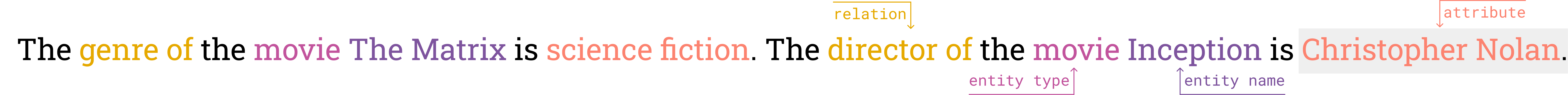}
    \vspace{-4mm}
\end{center}

Few-shot prompting improves generalization but remains brittle to surface-level changes, highlighting the need to assess robustness \citep{sclar2024quantifying}.

{\bf Random Statement.}
Prepend a fixed, unrelated grammatically correct sentence to the input prompt: ``The cat darted under the couch as the thunder cracked outside.''
\begin{center}
\vspace{-4mm}
    \includegraphics[width=.95\linewidth]{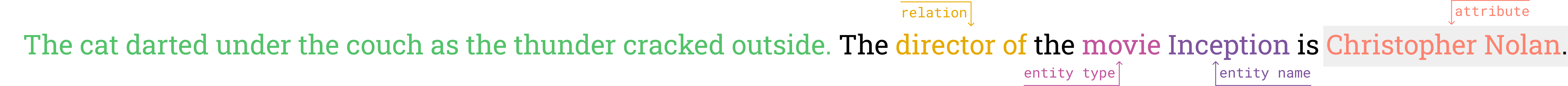}
    \vspace{-4mm}
\end{center}

Semantically neutral distractors, such as irrelevant prefixes, significantly affect model predictions, indicating sensitivity to prompt framing beyond content \citep{sclar2024quantifying}.

We assess the robustness of self-awareness directions to contextual perturbations at test time using the Gemma 2 2B model, shown in \autoref{tab:robust_modification_types}.
The model is trained on a fixed dataset, and its performance is evaluated across various input modifications.

\begin{table}[!t]
\caption{
Self-awareness directions robustness against various context perturbations for Gemma 2 2B model.
$\Delta$ indicates test accuracy gain over the random baseline.).
Standard deviation in parentheses.
}
\centering
\resizebox{\textwidth}{!}{%
\begin{tabular}{l|ccc|ccc}
\toprule
\textbf{Modification Type} & \multicolumn{3}{c|}{\textbf{Train (shared across all)}} & \multicolumn{3}{c}{\textbf{Test (varies by modification)}} \\
\cmidrule(lr){2-4} \cmidrule(lr){5-7}
& \textbf{Loss} & \textbf{AUC ROC} & \textbf{Accuracy} & \textbf{Loss} & \textbf{AUC ROC} & \textbf{Accuracy} \\
\midrule
\multirow{7}{*}{
\begin{tabular}[c]{@{}l@{}}
None \\
Quotation Marks (single) \\
Quotation Marks (double) \\
Statement Question \\
Few-Shot (Only) \\
Few-Shot (Unique) \\
Random Sentence
\end{tabular}
}
& \multirow{7}{*}{0.383 (0.063)} & \multirow{7}{*}{0.901 (0.061)} & \multirow{7}{*}{0.833 (0.052)} 
& \textbf{0.397} (0.064) & \textbf{0.896} (0.060) & \textbf{0.820} (0.056) \\
& & & & \textbf{0.431} (0.066) & \textbf{0.882} (0.057) & \textbf{0.802}(0.060) \\
& & & & 0.448 (0.066) & 0.877 (0.056) & 0.797 (0.072) \\
& & & & 0.520 (0.092) & 0.856 (0.056) & 0.756 (0.061) \\
& & & & 0.708 (0.183) & 0.771 (0.069) & 0.650 (0.083) \\
& & & & 0.538 (0.157) & 0.845 (0.074) & 0.753 (0.065) \\
& & & & 0.458 (0.070) & 0.871 (0.057) & 0.791 (0.061) \\
\bottomrule
\end{tabular}
}
\vspace{-4mm}
\label{tab:robust_modification_types}
\end{table}

Adding quotation marks—either single or double—yields only minor reductions in test accuracy (0.802 and 0.797) compared to the unmodified baseline (0.820), indicating robustness to superficial punctuation changes. Rephrasing factual prompts as questions (\textit{Statement Question}) leads to a larger performance drop (0.756), suggesting sensitivity to structural semantics beyond lexical form. In contrast, appending unrelated content (\textit{Random Sentence}) minimally affects performance (0.791), indicating resilience to distractors when the core entity–relation structure is preserved. The \textit{Few-Shot (Only)} setting causes substantial degradation (0.650), likely due to signal dilution, whereas \textit{Few-Shot (Unique)} better maintains accuracy (0.753), underscoring the role of relational diversity in preserving linear decodability.

Overall, these findings underscore that factual self-awareness in LMs is relatively robust to superficial noise but sensitive to semantically meaningful shifts in input structure.

\subsection{Impact of Sampling Parameters k-l on Probe Behavior}
\label{subsec:k-l-impact}

\begin{figure}[htbp]
    \centering

    \begin{minipage}[t]{0.48\textwidth}
        \centering
        \includegraphics[width=\linewidth]{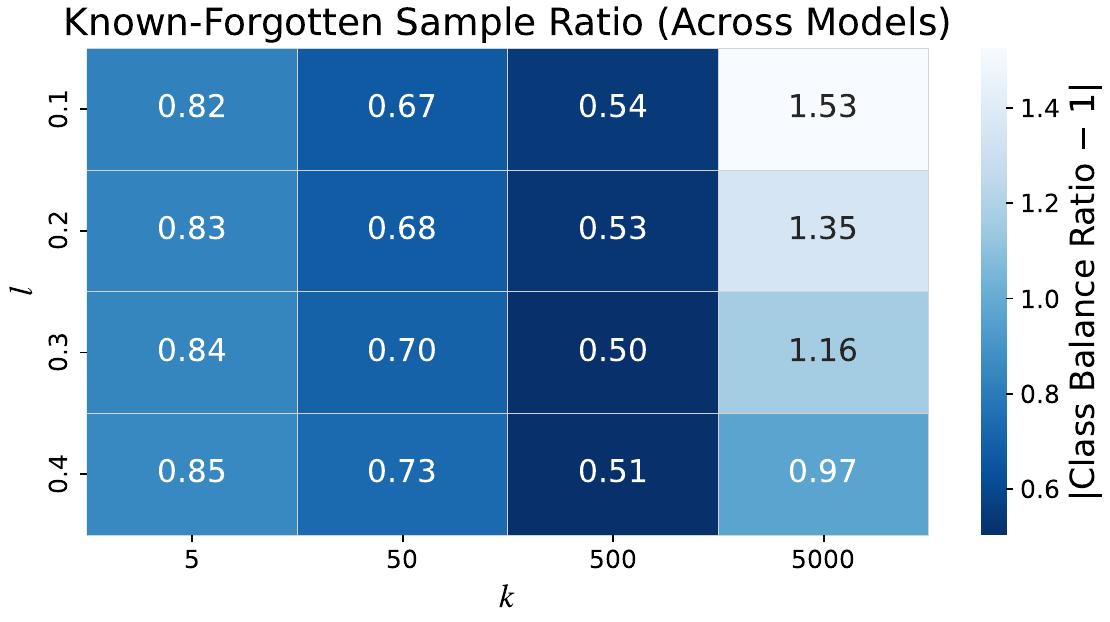}
        \caption{
            Known-Forgotten sample ratio for each $(k, l)$ configuration, aggregated across all models. 
            Lower values (darker) indicate more balanced retention, helping identify the globally optimal $(k, l)$ setting that generalizes across models.
        }
        \label{fig:gap-heatmap}
    \end{minipage}
    \hfill
    \begin{minipage}[t]{0.48\textwidth}
        \centering
        \includegraphics[width=\linewidth]{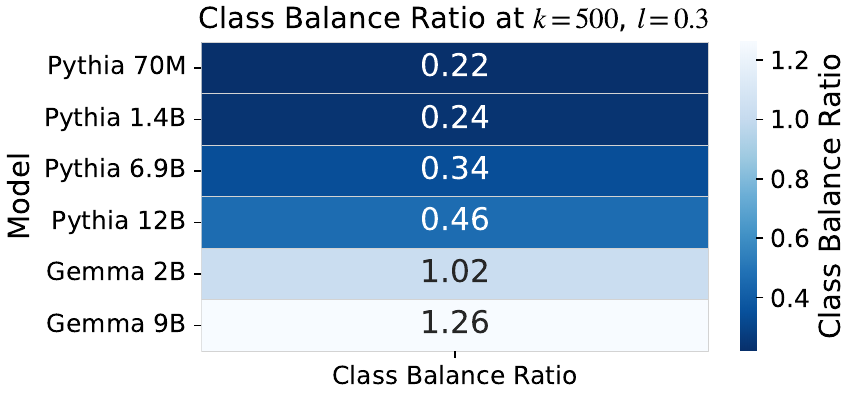}
\caption{
Class balance ratio at $k=500$, $l=0.3$ for each model. 
Values closer to 1.0 indicate more balanced retention, though Gemma 2B diverges significantly, suggesting this configuration may not generalize well to all architectures.
}
        \label{fig:k500-l01}
    \end{minipage}

\end{figure}

We systematically evaluate the effect of varying $k$ and $l$ values on linear probe performance and the class balance between \texttt{known} and \texttt{forgotten} samples for Gemma 2 (2B, 9B)~\citep{team2024gemma} and Pythia (70M, 1.4B, 6.9B, 12B)~\citep{biderman2023pythia} models.
Known-to-forgotten class ratios for different $k$–$l$ configurations per model $m$ are listed in \autoref{appendix:k-l_ratio_class_balance}, with visualizations in Figures~\ref{fig:class-balance-pythia} and~\ref{fig:class-balance-gemma} for Pythia and Gemma 2, respectively.
Among the Pythia models, Pythia-12B exhibits the highest known-to-forgotten ratio under the default $k$–$l$ setting, with $\text{Ratio}_{\text{Pythia-12B}}(k=500,\,l=0.3) = 0.46$.
In the Gemma 2 series, Gemma-2B shows a substantially more balanced attribution of factual knowledge, with $\text{Ratio}_{\text{Gemma-2B}}(k=500,\,l=0.3) = 1.02$.

To complement the analysis of class balance, we examine how the $(k, l)$ configuration affects downstream recovery of factual self-awareness in terms of linear probe performance.
In \autoref{fig:acc_gain_k_l_dep} in \autoref{appendix:k-l_ratio_class_balance}, we report the test and train accuracy improvements over a random baseline for the two representative models—Gemma 2B and Pythia 12B—across all evaluated $(k, l)$ settings.
These heatmaps reveal how performance varies with sampling parameters, where each cell shows the absolute accuracy gain (test/train) above random guessing.

Notably, linear probes on the Gemma 2B model exhibit consistent and substantial gains across multiple configurations, particularly around $k = 500$, whereas Pythia 12B shows smaller but more stable improvements. These results highlight the trade-off between class balance and discriminative performance, further motivating the choice of $(k = 500,\, l = 0.3)$ as a configuration that yields competitive accuracy.

\section{Scaling Behavior}
\label{sec:scaling_laws}

To further investigate the emergence of self-awareness directions in language models, we analyze models from the Pythia scaling suite \cite{biderman2023pythia}, focusing on variation across model sizes—specifically 70M, 1.4B, 6.9B, and 12B parameters.

\begin{figure}[ht]
    \centering
    \begin{minipage}[t]{0.44\linewidth}
        \centering
        \includegraphics[width=\linewidth]{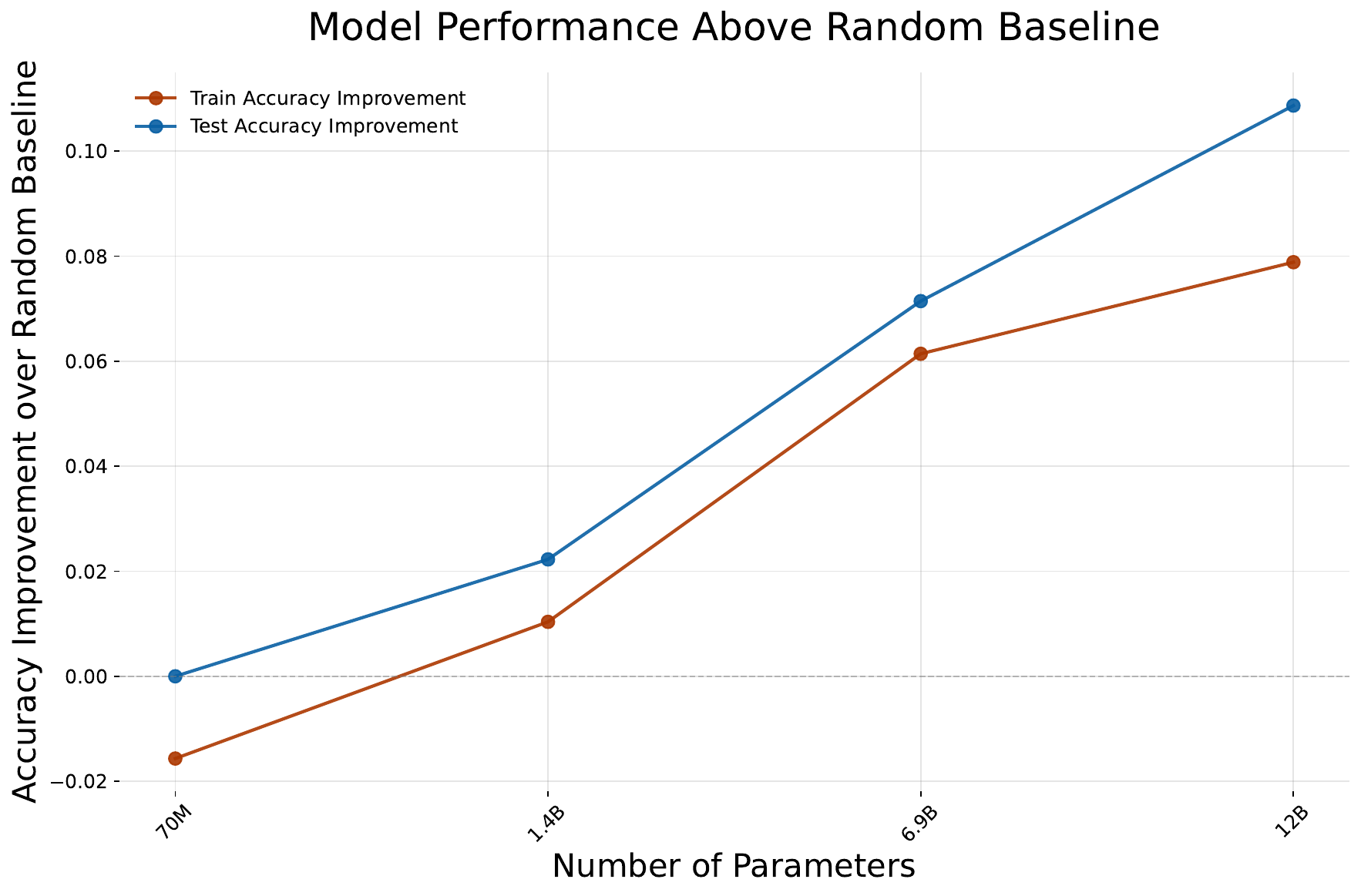}
        \caption{Accuracy gain over random baseline for training (orange) and test (blue) linear probes as a function of model size. Larger models exhibit greater gains in accuracy, with test performance benefiting more substantially from scaling.}
        \label{fig:scaling_laws_models}
    \end{minipage}
    \hfill
    \begin{minipage}[t]{0.52\linewidth}
        \centering
        \includegraphics[width=\linewidth]{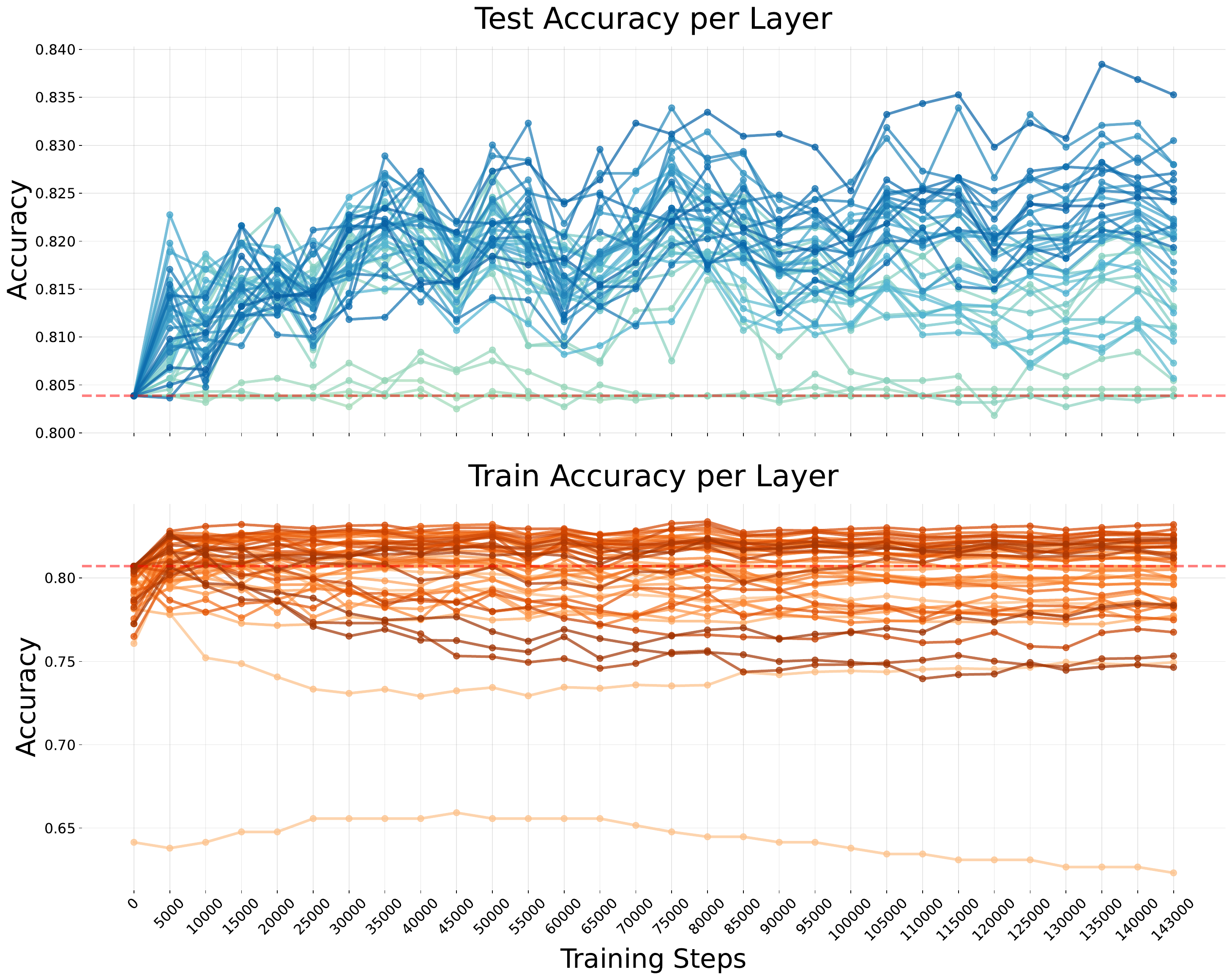}
        \caption{Linear probe accuracy across Pythia 1.4B training checkpoints/tokens. (Top) Training accuracy by layer (warmer colors = deeper layers). (Bottom) Test accuracy by layer (cooler colors = deeper layers, brighter = later layers). Red dashed line: random baseline.}
        \label{fig:accuracy_per_checkpoint_all_layers_1.4b}
    \end{minipage}
\end{figure}

We measure the emergence of linearly decodable features associated with self-awareness by computing the accuracy improvement of linear probes over a random baseline across models of increasing scale.
As shown in \autoref{fig:scaling_laws_models}, both training and test accuracy improvements grow monotonically with model size, indicating that larger models develop more robust and generalizable representations.
Notably, the improvement is more pronounced on the test set, suggesting that increased capacity enhances the transferability of these features beyond the training distribution.

To further examine how these features evolve during training, we evaluate checkpoints of Pythia 1.4B, spaced every 5000 steps from 0 to 143000, as shown in \autoref{fig:accuracy_per_checkpoint_all_layers_1.4b}.
At initialization (\texttt{0}), linear probe accuracy is at the random baseline, indicating no self-awareness directions in the untrained model.
During training, middle layers consistently yield the highest probe accuracy, while early and late layers perform worse.
Training accuracy rises quickly and plateaus early, while test accuracy improves more gradually.
On the test set, the highest accuracy occurs in the upper layers, suggesting self-awareness features are more strongly encoded later in the network.
In contrast, middle layers dominate on the training set, indicating a divergence in the distribution of generalizable vs. task-specific features across depth.

\section{Conclusion}
In this work, we explore the landscape of factual self-awareness of pretrained LMs. We precisely ask the question whether an LM encodes its certainty within its neural representations that it will be able to recall a given factual association.
We frame this as awareness-at-generation.
Providing an affirmative answer, we show that encoding of such a signal is linear and surprisingly robust against context perturbation.
We find that while a threshold model size is essential for the self-awareness signal to appear, the strength of the signal is not directly proportional to scaling.
Same trend is observed in terms of training scale: the signal appears quite early in training and saturates quickly.
We argue that this specific type of self-awareness, which is evident at the time of generation, can serve as a stronger entry point to curb LM hallucination compared to post-hoc truthfulness, investigated hitherto.
Compared to the latter, this awareness-at-generation can be repurposed to restrain the model from a generation attempt before the actual generation.

\section{Acknowledgments}
We thank Jingcheng Niu, Federico Tiblias and Alireza Bayat Makou for their feedback on an early draft of this work.

This research work has been funded by the German Federal Ministry of Education and Research and the Hessian Ministry of Higher Education, Research, Science and the Arts within their joint support of the National Research Center for Applied Cybersecurity ATHENE.

\section*{Limitations}
By definition, the investigated self-awareness signal is limited to factual recall alone.
There are multiple other forms of self-awareness that this work does not address.
We look into the rudimentary form of factual recall where all the necessary information (e.g., entity name, entity type, relation) is provided within the immediate query.
In open-ended generation tasks, the LM might need to gather this information from scattered context, resolve coreferences, perform multi-hop factual recall implicitly, etc.
We leave the investigation of self-awareness under such stressors as a future work.
Additionally, the dataset used in this study is restricted in its coverage of entity types and relation categories.
Expanding the dataset to include a broader and more diverse range of entities and relational structures would provide a more comprehensive understanding of how self-awareness representations generalize across semantic domains.
Finally, while we demonstrate the linearly separable signals of factual self-awareness in the intermediate neural representations and its scaling behavior, it is not known how the model learns to encode this from mere next token prediction training, or the internal causal components that construct and use these signals.

%%%%%%%%%%%%%%%%%%%%%%%%%%%%%%%%%%%%%%%%%%%%%%%%%%%%%%%%%%%%

\appendix
\section{Input Templates}
\label{appendix:data_template}

The selected relations include: \texttt{player} — `birth place', `birth date', `position', `nationality'; \texttt{movie} — `director', `release date', `genre', `country'; \texttt{city} — `country', `first mayor', `founded date', `climate type'; and \texttt{song} — `artist', `album', `release date', `language'.

Initially, we constructed the input templates using the relations described in \citet{ferrando2024do} for four categories: \texttt{football player}, \texttt{movie}, \texttt{city}, and \texttt{song}, we call this set of relations as \texttt{relations1}.
The specific relationships extracted for each category are as follows:
\begin{itemize}
  \item \texttt{player}: birthplace, birth date, teams played.
  \item \texttt{movie}: director, screenwriter, release date, genre, duration, cast.
  \item \texttt{city}: country, population, elevation, coordinates.
  \item \texttt{song}: artist, album involvement, publication year, genre.
\end{itemize}
Since the number of relations is not balanced for the categories and some relation answers have non-trivial modality, e.g. coordinates of a city, we propose a new unified dataset, that is balanced in the sense of number of features and standard output modalities, we call this set of relations \texttt{relations2}.
The following relations shape the \texttt{unified} dataset:
\begin{itemize}
  \item \texttt{player}: birthplace, birthdate, position, nationality.
  \item \texttt{movie}: director, release date, genre, production country.
  \item \texttt{city}: country, population, founded date, timezone.
  \item \texttt{song}: artist, album label, release date, language.
\end{itemize}

Afterwards, we hand-craft templates from the quadruples to form statements.
For example, "The movie Inception was directed by \textit{director} Christopher Nolan".
Since the expected answer for some relations can be ambiguous as in "The player Michael Jordan was born in \textit{city of}..." where it is unclear whether the response should be a location or a year—we incorporate "hints" at the end of each relation.

We experimented with four input templates (see \autoref{tab:input_templates}) using the \texttt{relations2} set, aiming to eliminate spurious correlations and isolate the self-awareness signal captured by linear probes.
Notably, only \texttt{template2} consistently captures the self-awareness signal without interference from confounding factors.

\begin{table}[H]
    \centering
    \begin{tabular}{ll}
        \toprule
        \textbf{Template Name} & \textbf{Sample Sentence} \\
        \midrule
        \texttt{template1} & "The player Youri Djorkaeff was born in the city of" \\
        \texttt{template1\_const\_end} & "The player Youri Djorkaeff's birth city is" \\
        \texttt{template2} & "The city of birth for the player Youri Djorkaeff is" \\
        \texttt{template2\_balanced} & "The city of birth for the player Youri Djorkaeff is" \\
        \bottomrule
    \end{tabular}
    \caption{Input templates using the entity type "player" and entit name "Youri Djorkaeff"}
    \label{tab:input_templates}
\end{table}

\texttt{template1}: Places the entity type and name at the beginning of the sentence and ends with variable tokens. For example: \textit{``The \textless entity\_type\textgreater\ \textless entity\_name\textgreater\ was born in the city of...''}
    
\texttt{template1\_const\_end}: Similar to \texttt{template1}, it places the entity type and name at the beginning of the sentence but always ends with the fixed token \textit{"is"}. For example: \textit{``The \textless entity\_type\textgreater\ \textless entity\_name\textgreater's birth city is...''}
    
\texttt{template2}: In contrast to \texttt{template1}, it places the entity type and name at the end of the prompt. Like \texttt{template1\_const\_end}, it also ends with the token \textit{"is"}. For example: \textit{``The city of birth for the \textless entity\_type\textgreater\ \textless entity\_name\textgreater\ is...''}

\texttt{template2\_balanced}: Further improvement of \texttt{template2} to have balanced number of known and forgotten samples across relations per category.

See the full set of templates from \texttt{template2\_balanced} that were used in the experiments \autoref{tab:input_templates}.

\section{Sample Distribution Across Entity Types}
\label{appendix:dataset_distribution}

We present the full distribution of known and forgotten samples using \texttt{template2\_balanced} for Gemma 2 2B with $k=500$ and $l=0.3$ parameters in \autoref{tab:data_sample_distribution_2b} and for Pythia 12B in \autoref{tab:data_sample_distribution_12b}.

\begin{table}[H]
\centering
\captionsetup{font=small}
\begin{minipage}{0.48\textwidth}
\centering
\caption{Distribution of known and forgotten samples across entity categories for the Gemma 2 2B model, using top-$k=500$ and bottom-$l=0.3$ thresholds.}
\small
\begin{tabular}{lccc}
\toprule
\textbf{Category} & \textbf{Known} & \textbf{Forgotten} & \textbf{Subset Total} \\
\midrule
Player & 1286 & 2922 & 4208 \\
Movie  & 3602 & 1810 & 5412 \\
City   & 1017 & 541  & 1558 \\
Song   & 1475 & 1995 & 3470 \\
\midrule
\textbf{Total} & 7380 & 7268 & 14648 \\
\bottomrule
\end{tabular}
\label{tab:data_sample_distribution_2b}
\end{minipage}
\hfill
\begin{minipage}{0.48\textwidth}
\centering
\caption{Distribution of known and forgotten samples across entity categories for the Pythia 12B model, using top-$k=500$ and bottom-$l=0.3$ thresholds.}
\small
\begin{tabular}{lccc}
\toprule
\textbf{Category} & \textbf{Known} & \textbf{Forgotten} & \textbf{Subset Total} \\
\midrule
Player & 657  & 3551 & 4208 \\
Movie  & 3602 & 1810 & 5412 \\
City   & 574  & 984  & 1558 \\
Song   & 538  & 2932 & 3470 \\
\midrule
\textbf{Total} & 5371 & 9277 & 14648 \\
\bottomrule
\end{tabular}
\label{tab:data_sample_distribution_12b}
\end{minipage}
\end{table}

\section{Separation Scores}
\label{appendix:separation_scores}

\begin{figure}[H]
    \centering
    \includegraphics[width=0.48\linewidth]{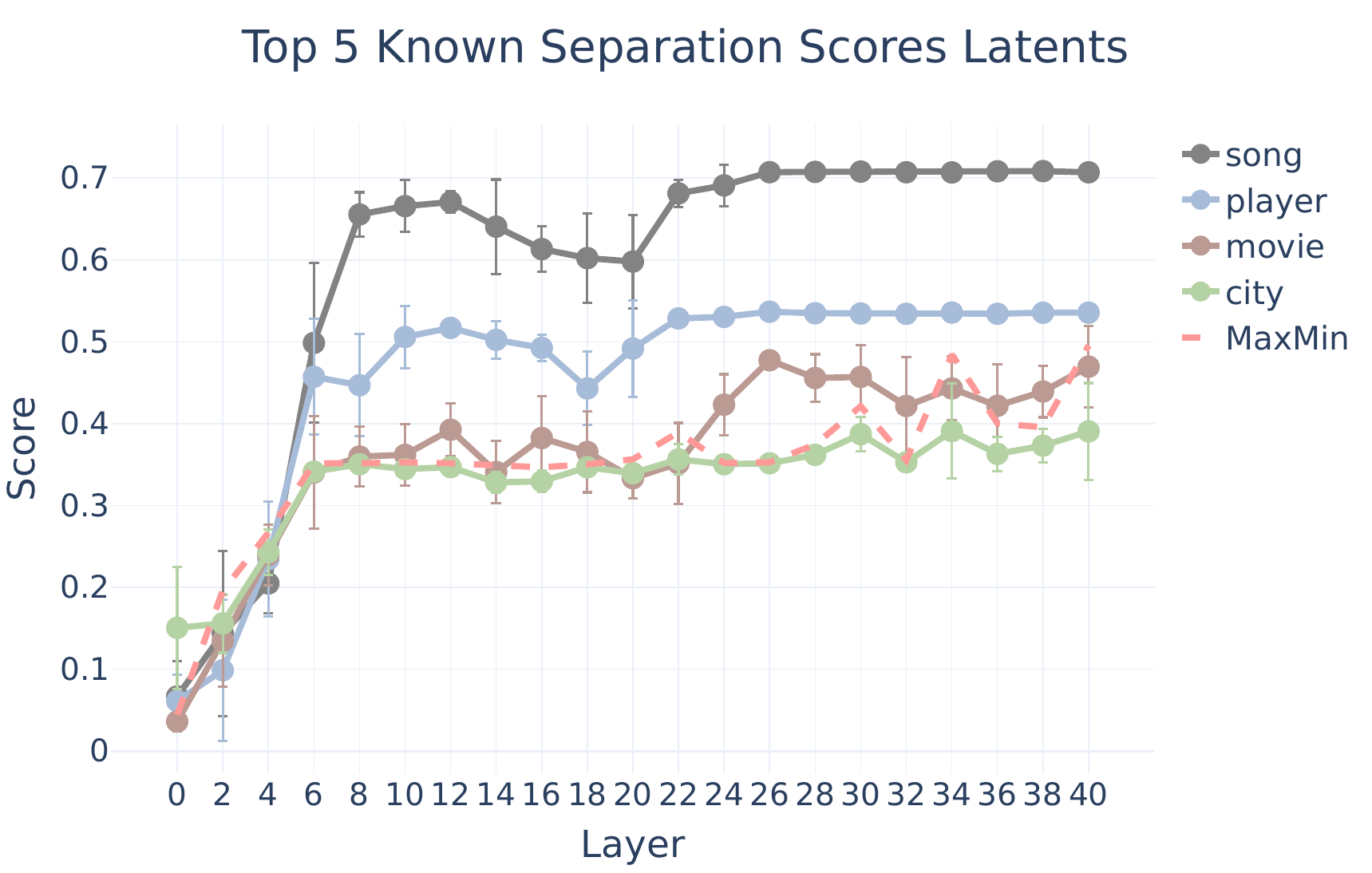}
    \hfill
    \includegraphics[width=0.48\linewidth]{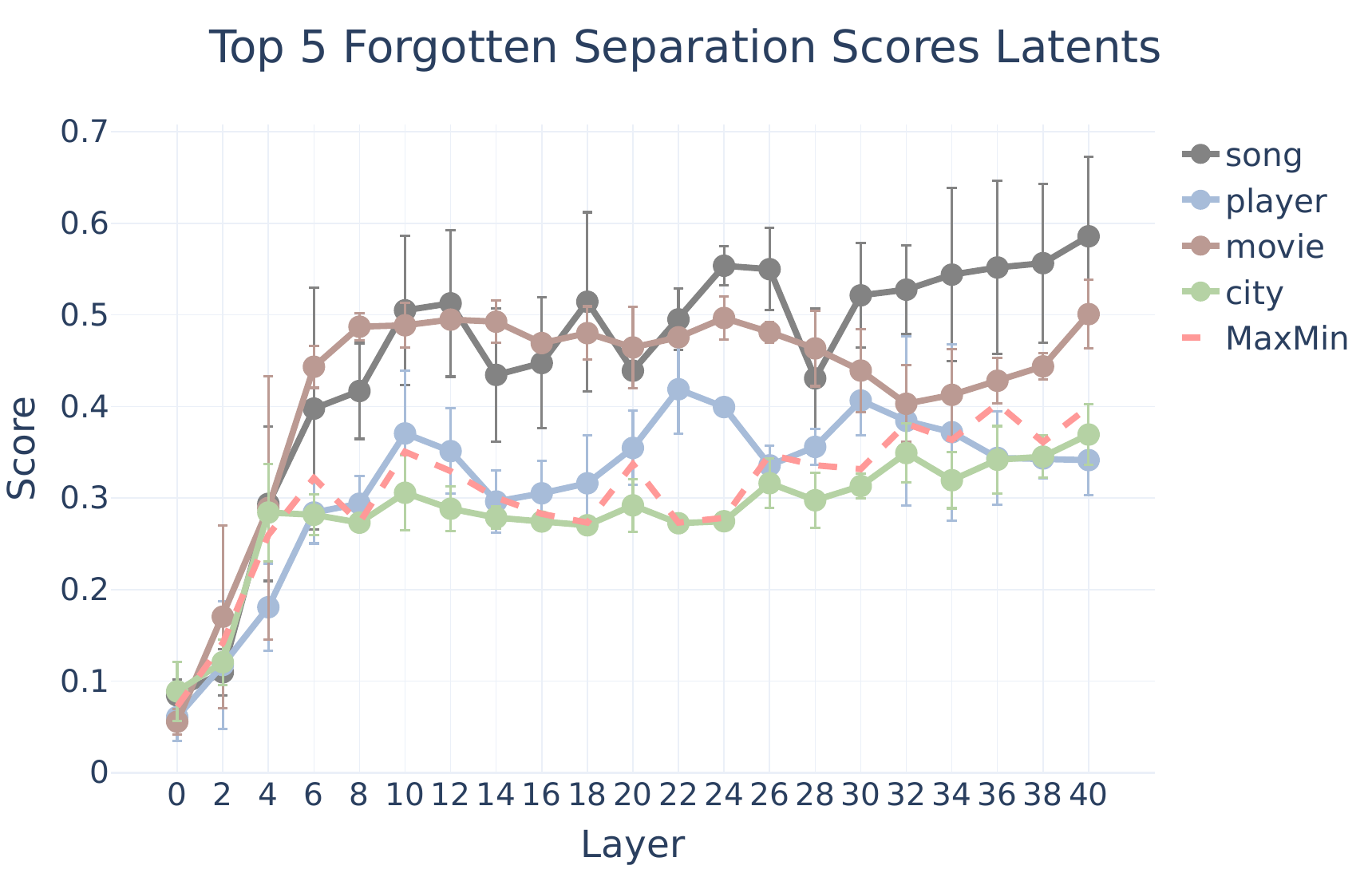}
    \caption{Latent separation scores using SAE activations on Gemma 2 9B. Left: \textbf{known} entities. Right: \textbf{forgotten} entities.}
    \label{fig:sae_latent_separation_comparison_gemma_9b}
\end{figure}

\begin{figure}[H]
    \centering
    \includegraphics[width=0.48\linewidth]{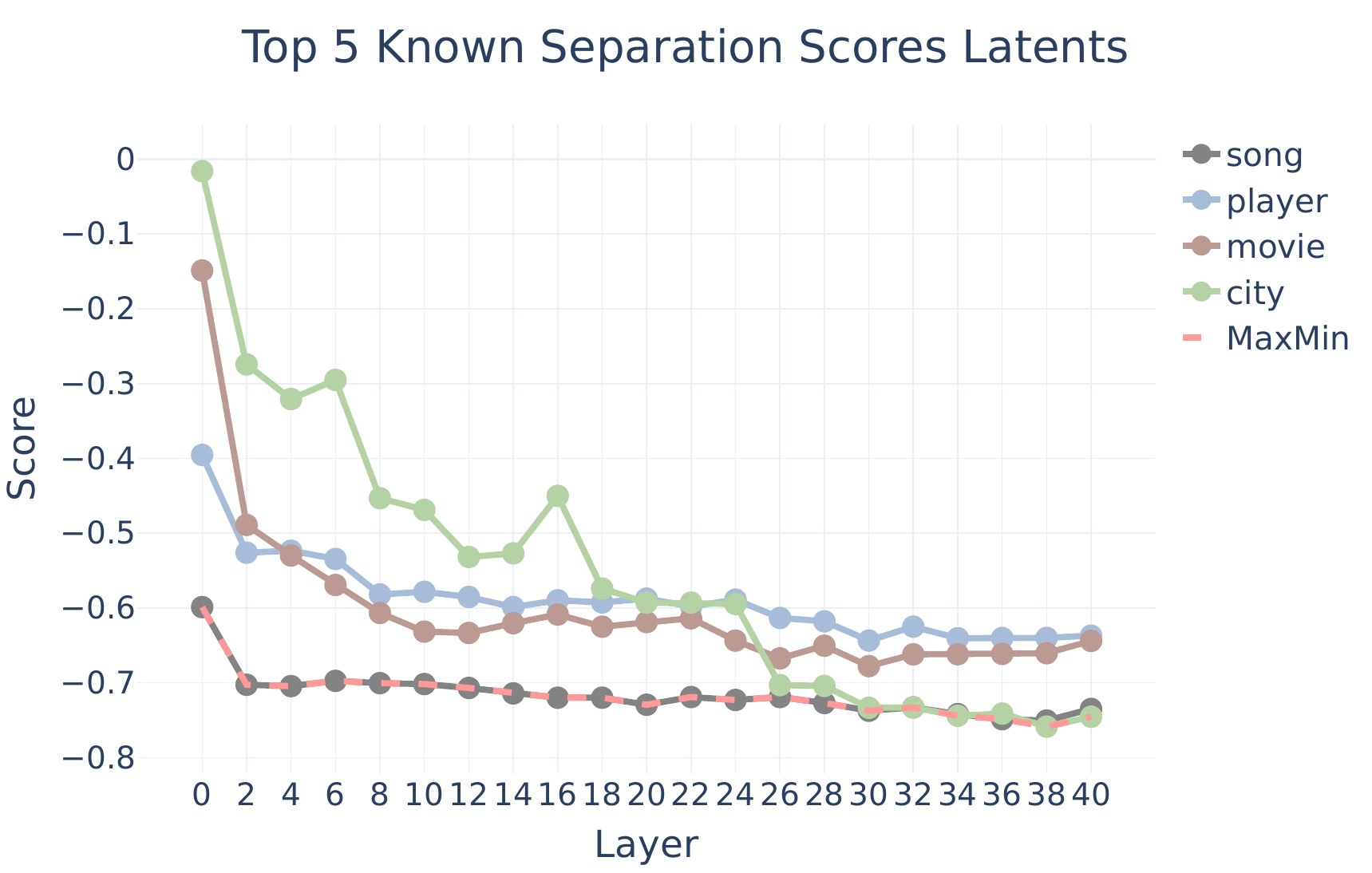}
    \hfill
    \includegraphics[width=0.48\linewidth]{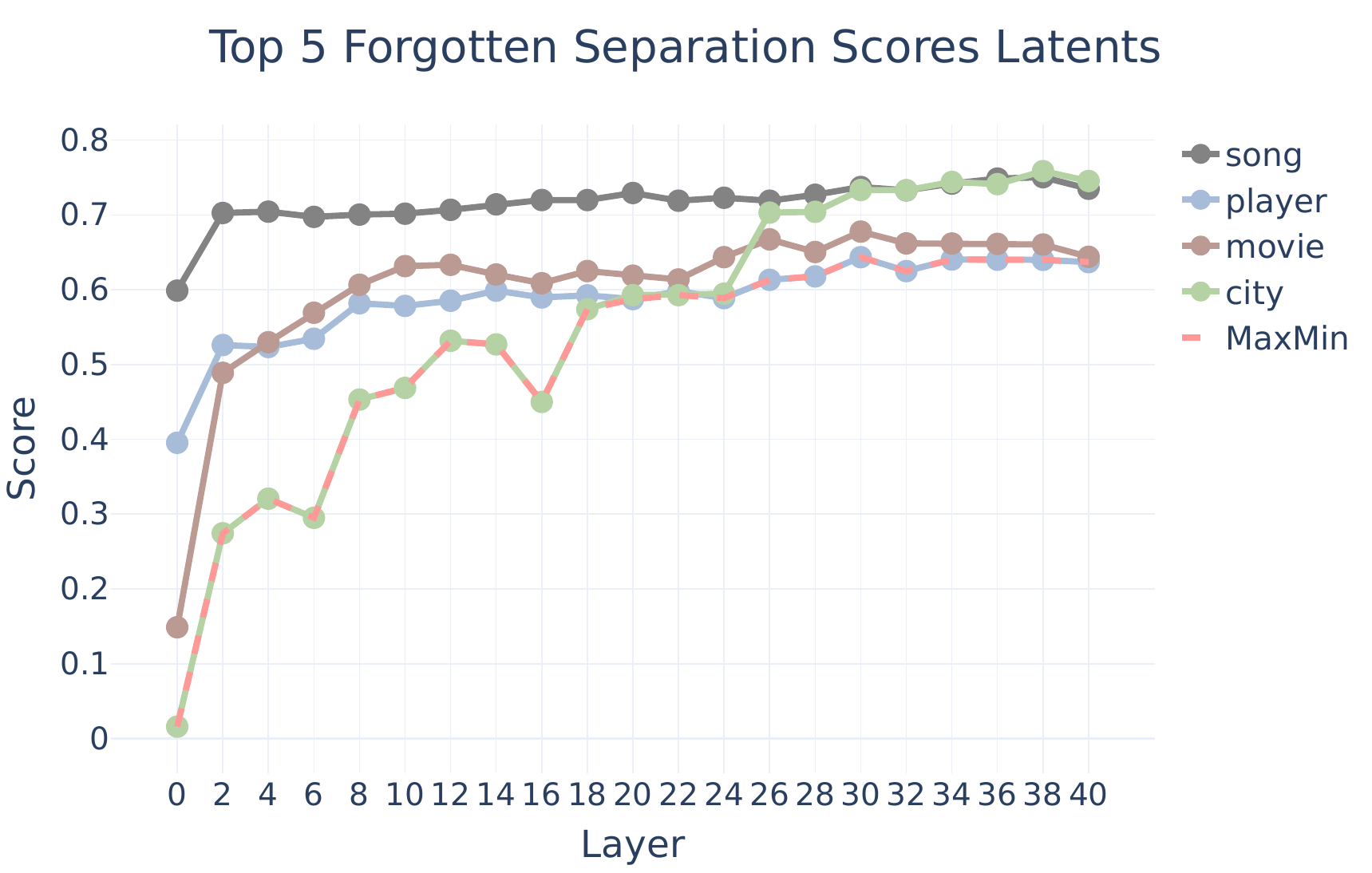}
    \caption{Latent separation scores using Linear Probe activations on Gemma 2 9B. Left: \textbf{known} entities. Right: \textbf{forgotten} entities.}
    \label{fig:linear_prob_latent_separation_comparison_gemma_9b}
\end{figure}

The latent separation scores for Gemma 2 9B using SAE activations (\autoref{fig:sae_latent_separation_comparison_gemma_9b}) and Linear Probe activations (\autoref{fig:linear_prob_latent_separation_comparison_gemma_9b}).

\section{(k,l) Pair Impact on Probe Behavior}
\label{appendix:k-l_ratio_class_balance}

\begin{figure}[htbp]
    \centering
    % First actual image
    \begin{minipage}[t]{0.44\textwidth}
        \centering
        \includegraphics[width=\linewidth]{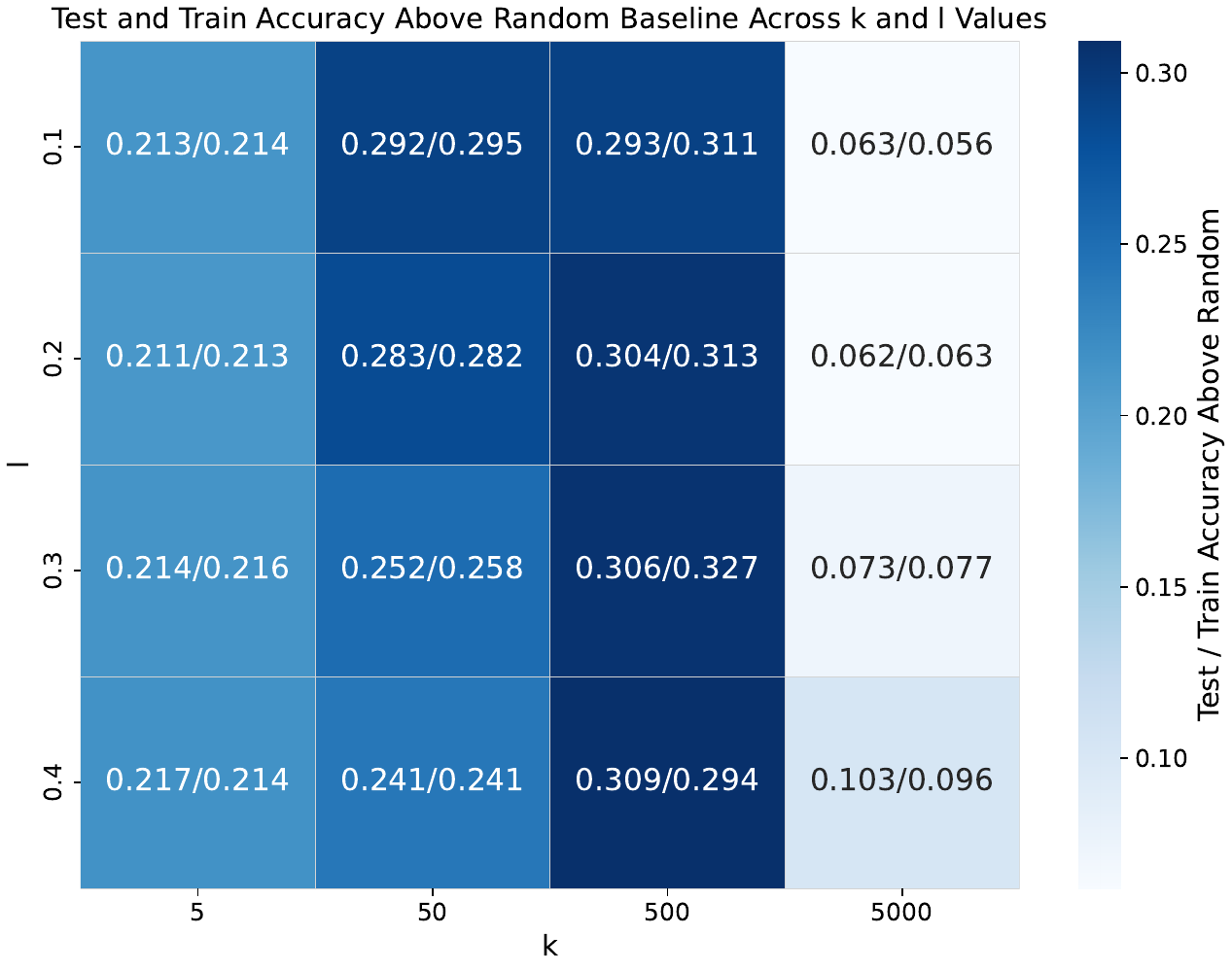}
        \caption*{(a)}
    \end{minipage}%
    \hfill
    % Second placeholder
    \begin{minipage}[t]{0.44\textwidth}
        \centering
        \includegraphics[width=\linewidth]{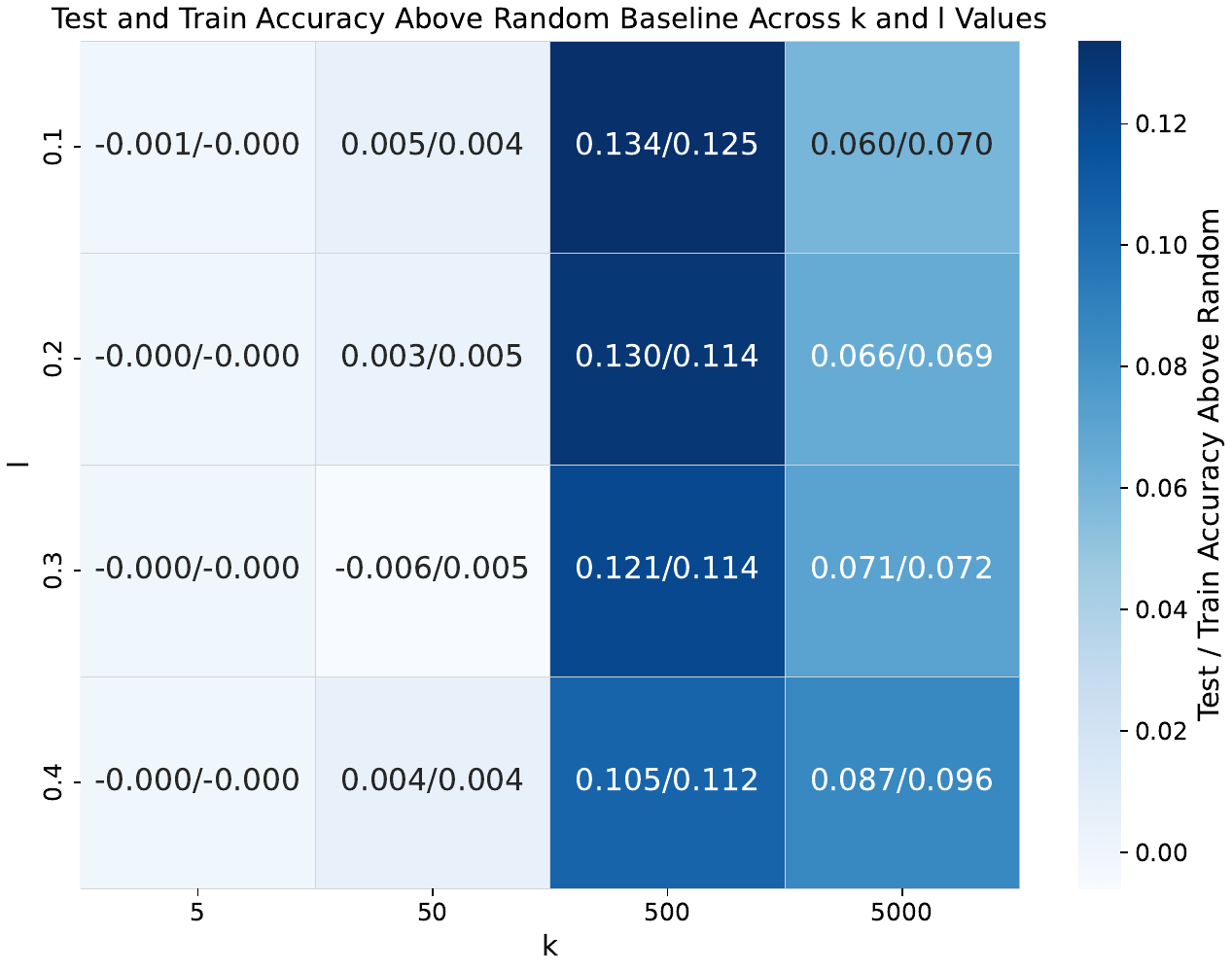}
        \caption*{(b)}
    \end{minipage}

    \caption{
    Accuracy gain over a random baseline from linear probes on: (a) Gemma 2 2B, and (b) Pythia 12B.
    Each cell displays the test/train accuracy above random for a given combination of $(k,l)$.
    Darker blue indicates greater accuracy gains.
    }
    \label{fig:acc_gain_k_l_dep}
\end{figure}

The figures in this section illustrate how varying the $(k, l)$ parameters—representing the number of known and forgotten samples, respectively—affects linear probe performance and class balance outcomes across different model scales. Figure~\ref{fig:acc_gain_k_l_dep} presents the test and train accuracy gains over a random baseline for the Gemma 2 2B and Pythia 12B models. Notably, we observe that increases in $k$ (the number of known samples) generally correspond to higher accuracy gains, particularly when $l$ (the number of forgotten samples) remains low. This trend is more pronounced in larger models, consistent with their greater capacity to capture and retain class-discriminative features.

\begin{figure}[htbp]
    \centering
    \begin{subfigure}{0.48\linewidth}
        \includegraphics[width=\linewidth]{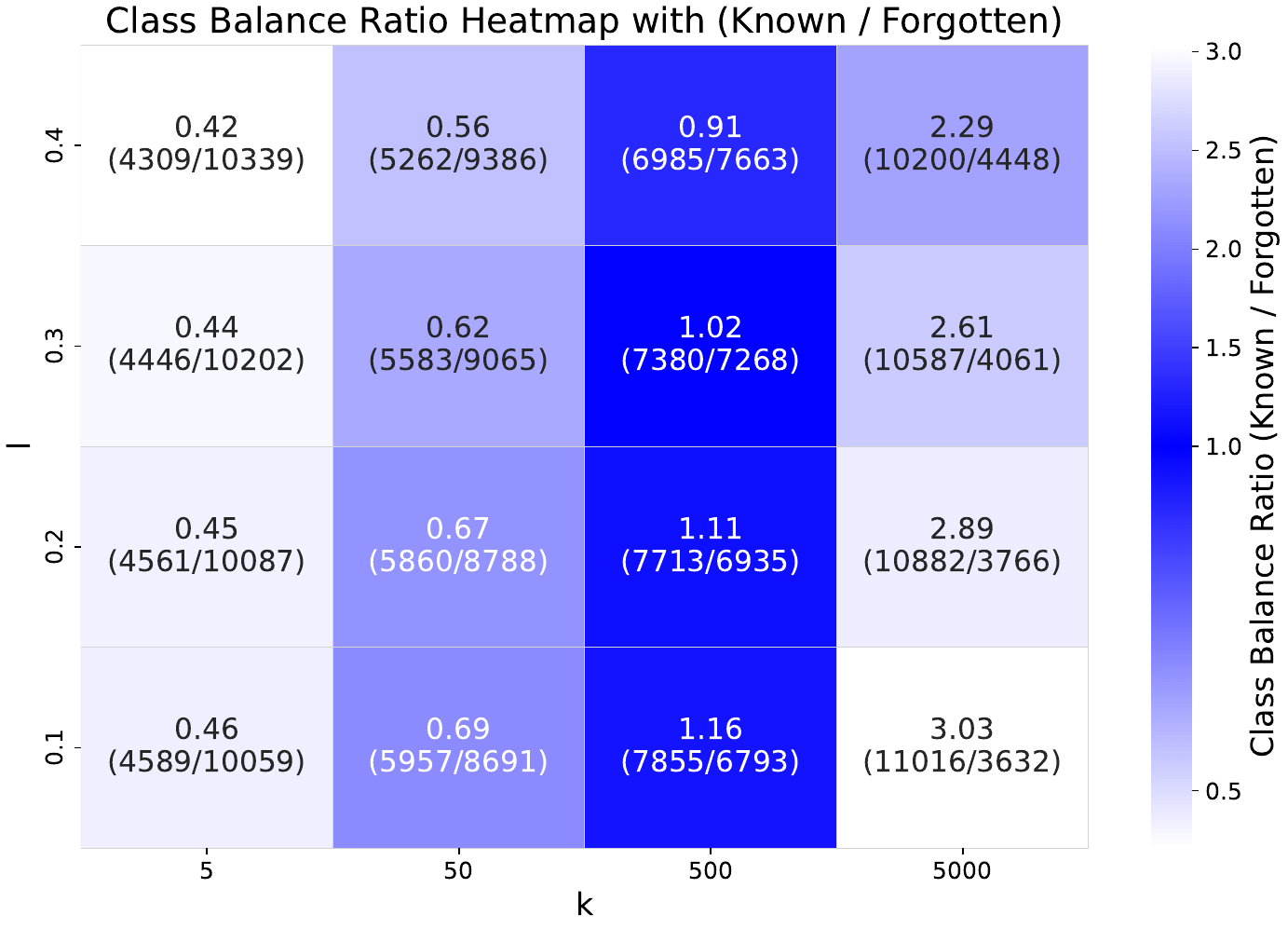}
        \caption{Gemma 2 2B}
    \end{subfigure}
    \hfill
    \begin{subfigure}{0.48\linewidth}
        \includegraphics[width=\linewidth]{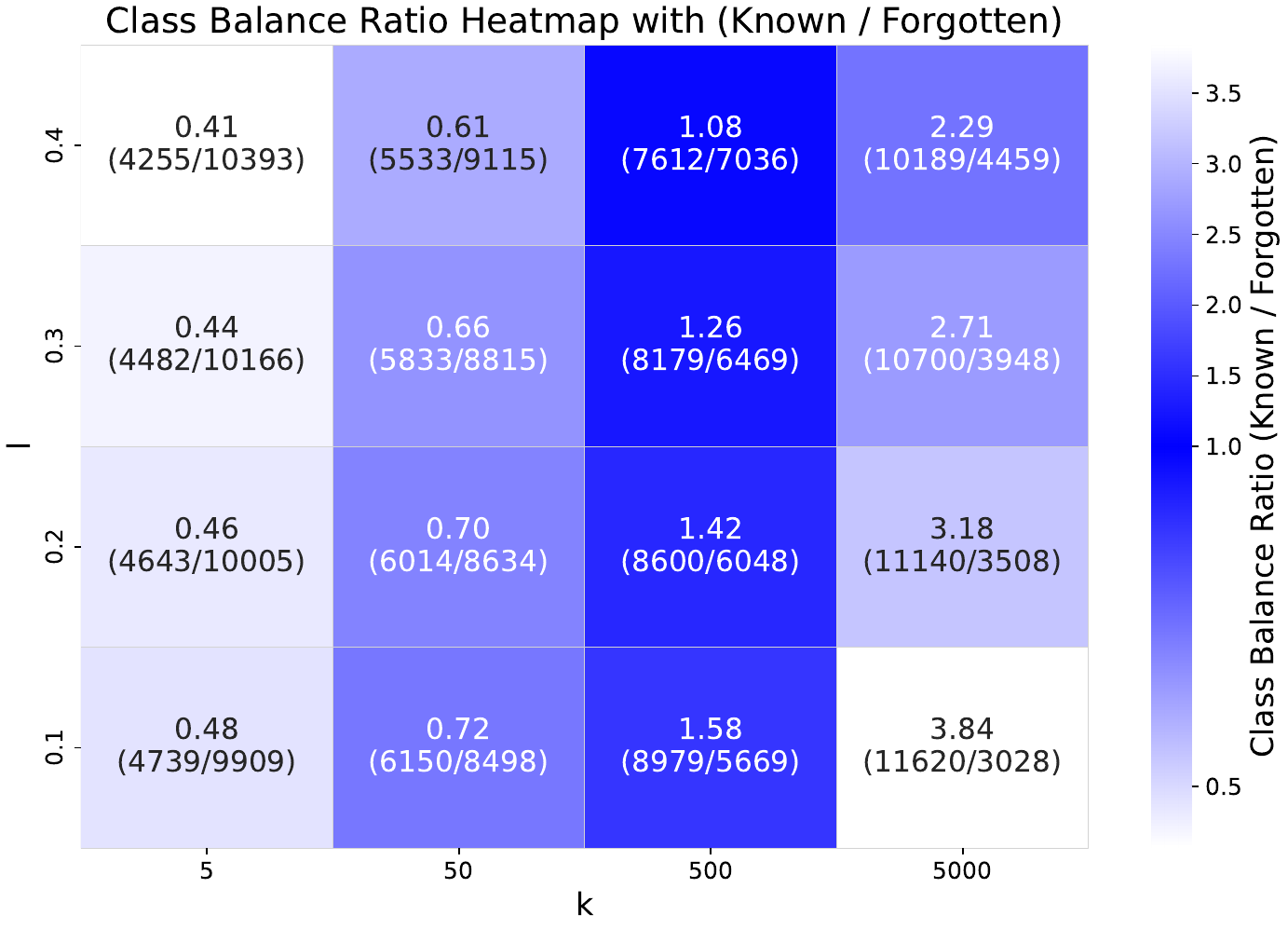}
        \caption{Gemma 2 9B}
    \end{subfigure}
    \caption{$k$-$l$ parameters dependence on number of class balance (known and forgotten samples ratio) heatmaps for Gemma 2 models.}
    \label{fig:class-balance-gemma}
\end{figure}

\begin{figure}[htbp]
    \centering
    % First row
    \begin{subfigure}{0.45\linewidth}
        \includegraphics[width=\linewidth]{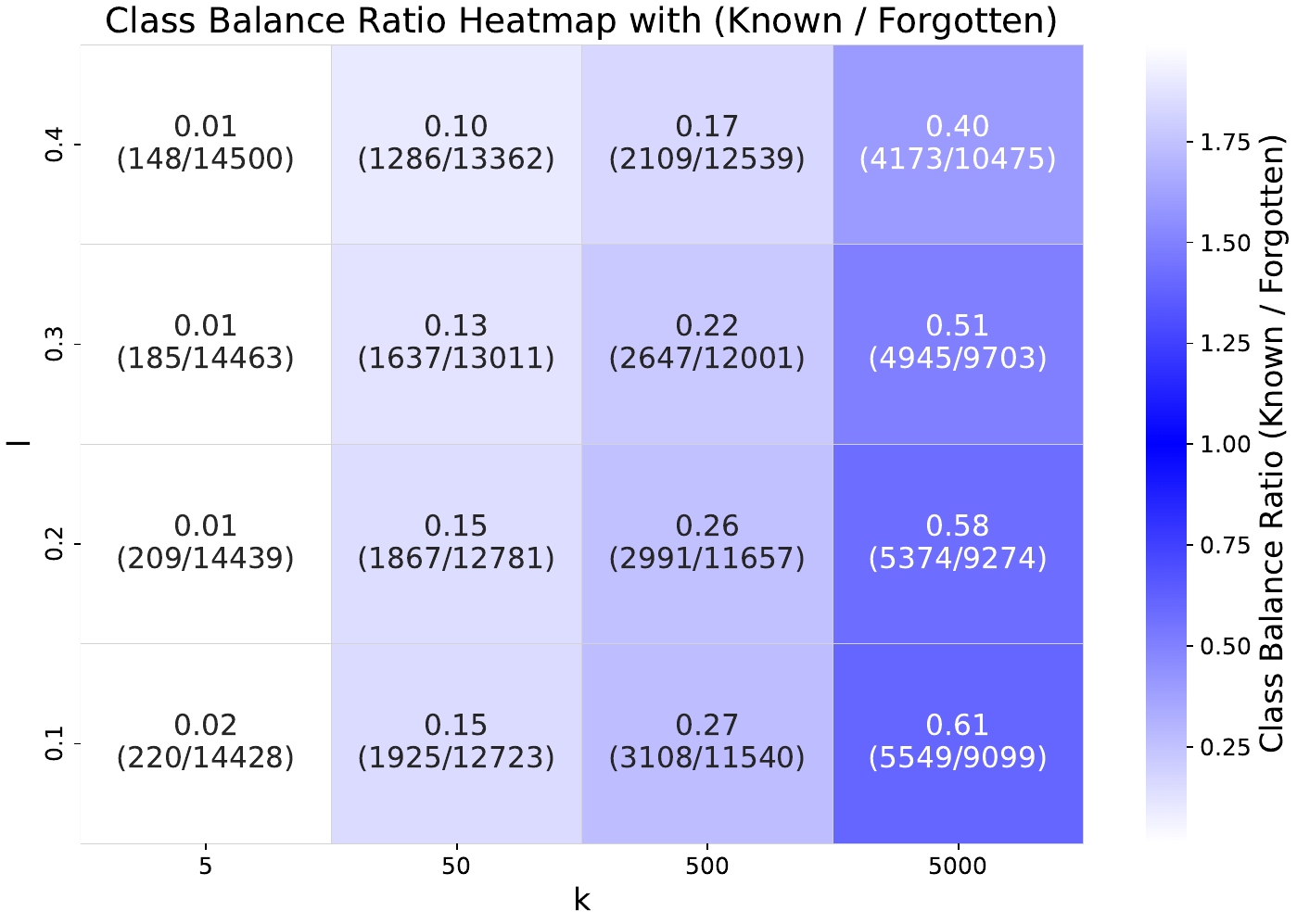}
        \caption{Pythia 70M}
    \end{subfigure}
    \hfill
    \begin{subfigure}{0.45\linewidth}
        \includegraphics[width=\linewidth]{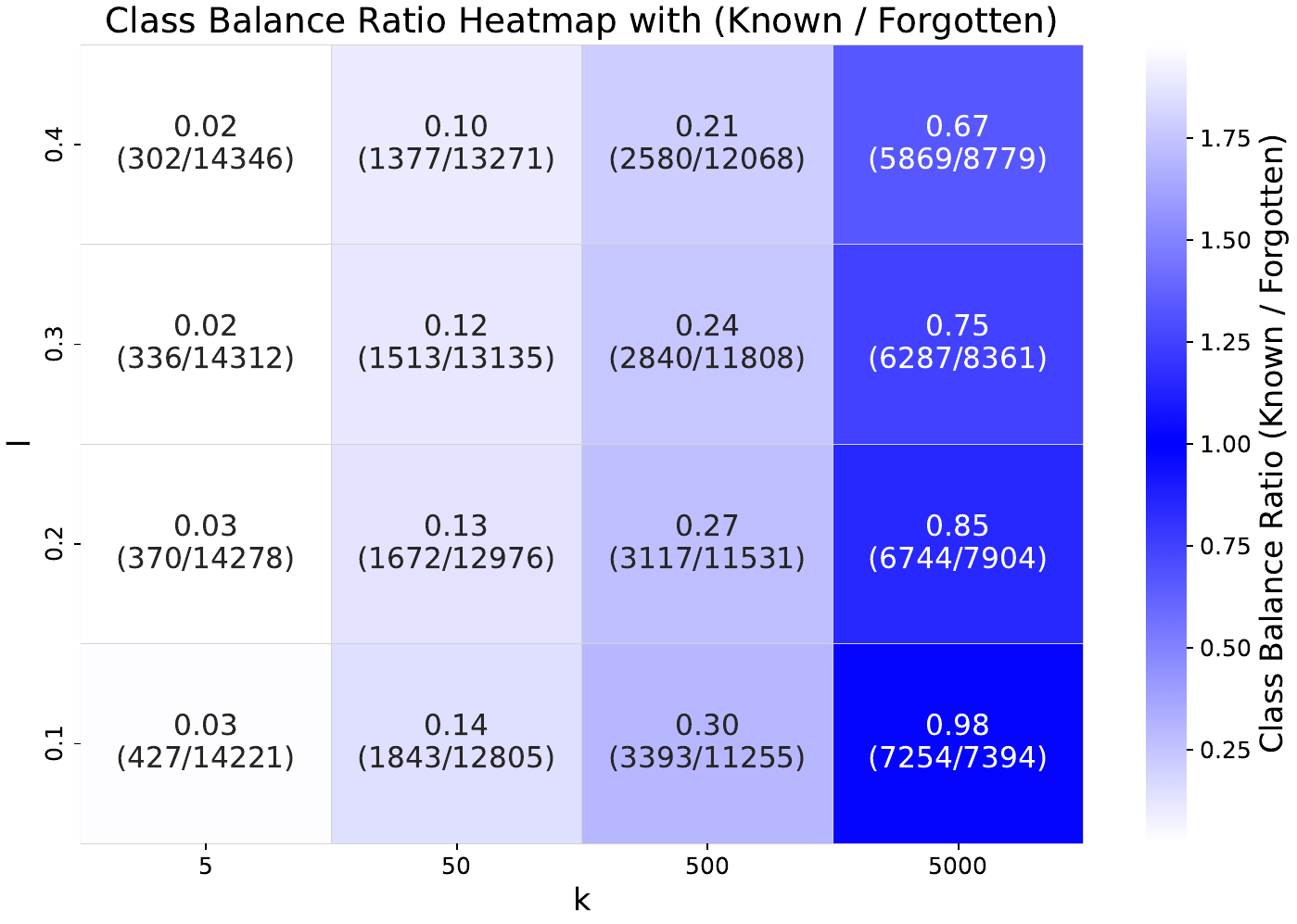}
        \caption{Pythia 1.4B}
    \end{subfigure}

    \vspace{0.5em}  % vertical spacing between rows

    % Second row
    \begin{subfigure}{0.45\linewidth}
        \includegraphics[width=\linewidth]{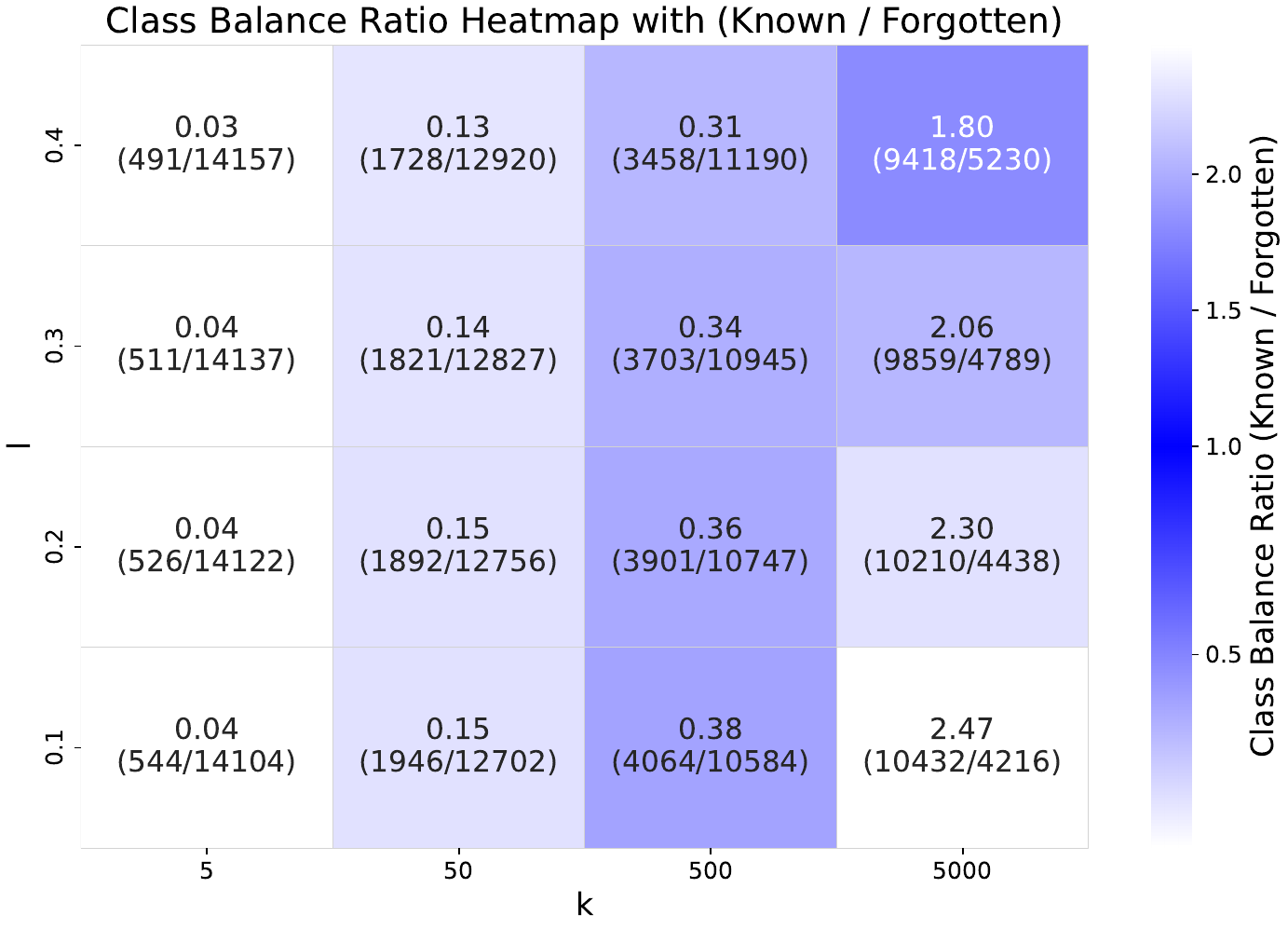}
        \caption{Pythia 6.9B}
    \end{subfigure}
    \hfill
    \begin{subfigure}{0.45\linewidth}
        \includegraphics[width=\linewidth]{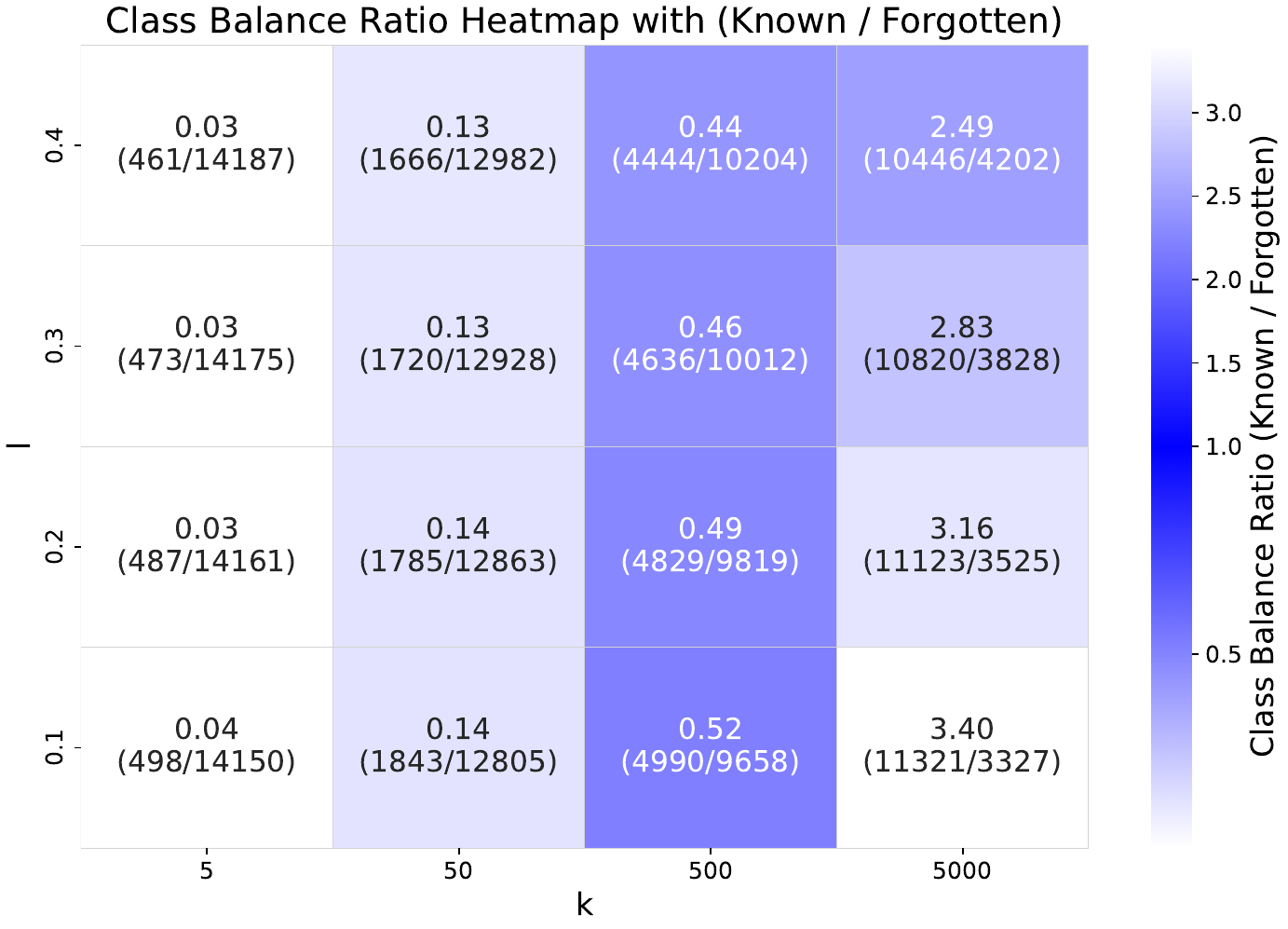}
        \caption{Pythia 12B}
    \end{subfigure}

    \caption{$k$-$l$ parameters dependence on number of class balance (known and forgotten samples ratio) heatmaps for Pythia models.}
    \label{fig:class-balance-pythia}
\end{figure}

Figures~\ref{fig:class-balance-pythia} and~\ref{fig:class-balance-gemma} explore the implications of $(k, l)$ settings on class balance, defined as the ratio of known to forgotten samples, for Pythia and Gemma 2 model families, respectively. The heatmaps indicate that this ratio grows with increasing $k$ and decreasing $l$, with larger models showing a more marked divergence between known and forgotten categories. These findings highlight the sensitivity of probe performance and interpretability metrics to sampling configurations, underscoring the importance of systematic calibration of $(k, l)$ pairs when designing probing protocols.

\end{document}